\documentclass[10pt,twocolumn,letterpaper]{article}
\usepackage[T1]{fontenc}
\usepackage{wacv}              % To produce the CAMERA-READY version
\usepackage[accsupp]{axessibility}
\usepackage{graphicx}
\usepackage{amsmath}
\usepackage{amssymb}
\usepackage{caption}
\usepackage{subcaption}
\newtheorem{definition}{Definition}
\usepackage{booktabs}
\usepackage{verbatim}
\usepackage[table]{xcolor}% http://ctan.org/pkg/xcolor
\usepackage[numbers,sort&compress]{natbib}
\usepackage{enumitem}
\usepackage{fvextra}
\makeatletter % changes the catcode of @ to 11
\newcommand{\linebreakand}{%
  \end{@IEEEauthorhalign}
  \hfill\mbox{}\par
  \mbox{}\hfill\begin{@IEEEauthorhalign}
}
\makeatother % changes the catcode of @ back to 12

\usepackage{float}
\usepackage[labelfont=bf,textfont=it]{caption}

\usepackage[pagebackref,breaklinks]{hyperref}

\hypersetup{colorlinks,
      linkcolor=blue,
      citecolor=blue,
      urlcolor=blue}
\newcommand{\cref}[2]{\hyperref[#2]{#1~\ref*{#2}}}
\newcommand{\colref}[2]{\hyperref[#2]{#1~\ref*{#2}}}

\newcommand{\figref}[1]{\colref{Figure}{#1}}

\newcommand{\secref}[1]{\colref{Section}{#1}}
\newcommand{\tabref}[1]{\colref{Table}{#1}}
\newcommand{\coloredref}[2]{\hyperref[#2]{#1~\ref*{#2}}}
\newcommand{\coloredsubref}[3]{\hyperref[#2]{#1~\ref*{#2}{#3}}}
\newcommand{\Algref}[1]{\hyperref[#1]{Algorithm~\ref*{#1}}}

\usepackage[capitalize]{cleveref}
\crefname{section}{Sec.}{Secs.}
\Crefname{section}{Section}{Sections}
\Crefname{table}{Table}{Tables}
\crefname{table}{Tab.}{Tabs.}

\def\wacvPaperID{2776} % *** Enter the WACV Paper ID here
\def\confName{WACV}
\def\confYear{2025}

\begin{document}

\title{Leveraging Vision Language Models for Specialized Agricultural Tasks}

% \and
\author{Muhammad Arbab Arshad$^1$ \quad Talukder Zaki Jubery$^1$ \quad Tirtho Roy$^1$ \quad Rim Nassiri$^1$ \and
Asheesh K. Singh$^1$ \quad Arti Singh$^1$ \quad Chinmay Hegde$^2$ \quad Baskar Ganapathysubramanian$^1$ \and
Aditya Balu$^1$ \quad Adarsh Krishnamurthy$^1$ \quad Soumik Sarkar$^{1,*}$\\[2ex]
\small $^1$Iowa State University, USA\\
\small $^2$New York University, USA\\
\small *Corresponding author: soumiks@iastate.edu}
\maketitle
\begin{figure*}[t!]
    \centering
    \includegraphics[width=0.99\linewidth,trim=0.0in 0.0in 0.0in 0.0in,clip]{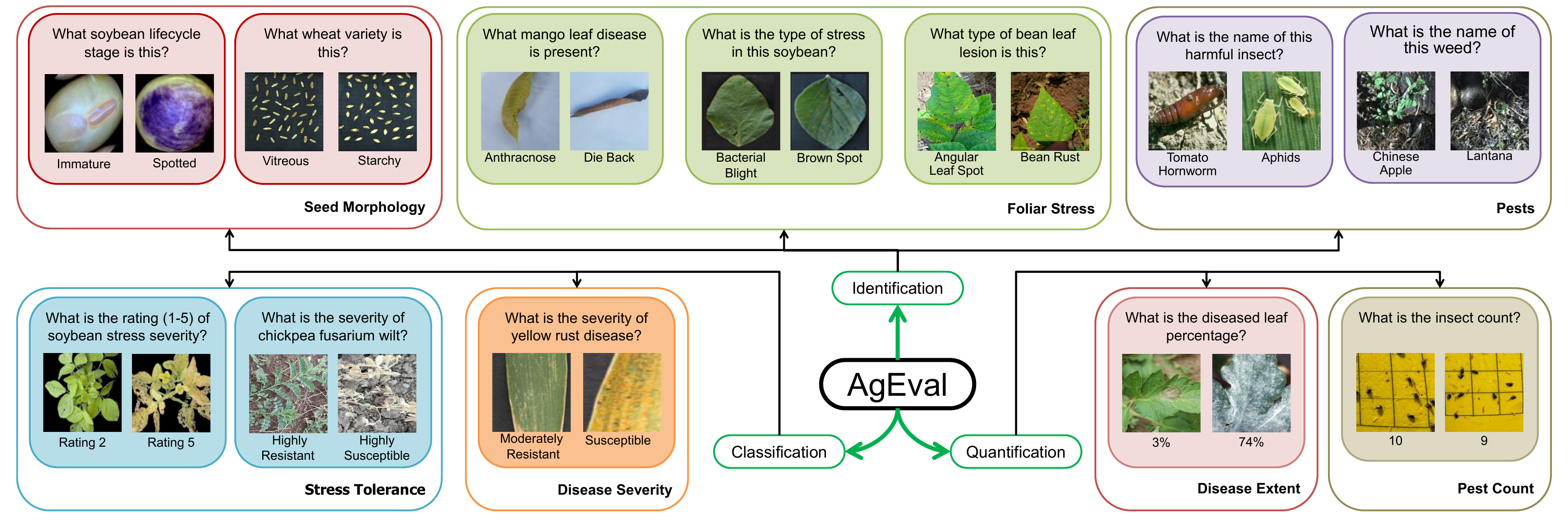}
    \caption{Overview of the AgEval benchmark. The figure showcases sample images across different types of tasks and specific problems, representing diverse plant stress phenotyping challenges in agriculture.}
    \label{fig:datasets}
\end{figure*}
%%%%%%%%% ABSTRACT
\begin{abstract}
As Vision Language Models (VLMs) become increasingly accessible to farmers and agricultural experts, there is a growing need to evaluate their potential in specialized tasks. We present AgEval, a comprehensive benchmark for assessing VLMs' capabilities in plant stress phenotyping, offering a solution to the challenge of limited annotated data in agriculture. Our study explores how general-purpose VLMs can be leveraged for domain-specific tasks with only a few annotated examples, providing insights into their behavior and adaptability. AgEval encompasses 12 diverse plant stress phenotyping tasks, evaluating zero-shot and few-shot in-context learning performance of state-of-the-art models including Claude, GPT, Gemini, and LLaVA. Our results demonstrate VLMs' rapid adaptability to specialized tasks, with the best-performing model showing an increase in F1 scores from 46.24\% to 73.37\% in 8-shot identification. To quantify performance disparities across classes, we introduce metrics such as the coefficient of variation (CV), revealing that VLMs' training impacts classes differently, with CV ranging from 26.02\% to 58.03\%. We also find that strategic example selection enhances model reliability, with exact category examples improving F1 scores by 15.38\% on average. AgEval establishes a framework for assessing VLMs in agricultural applications, offering valuable benchmarks for future evaluations. Our findings suggest that VLMs, with minimal few-shot examples, show promise as a viable alternative to traditional specialized models in plant stress phenotyping, while also highlighting areas for further refinement. Results and benchmark details are available at: \url{https://github.com/arbab-ml/AgEval}
\end{abstract}
%%%%%%%%% BODY TEXT
\section{Introduction}
\label{sec:intro}
Food security is a critical global challenge requiring sustainable improvements in agricultural productivity~\citep{anderson2004emerging}. Agricultural research increasingly utilizes computer vision and AI to optimize crop management and enhance yield, profitability, and sustainability~\citep{sarkar2023cyber}. Phenotyping involves visually inspecting plants to extract agronomically relevant features. While traditional methods are time-consuming and labor-intensive, recent advancements in computer vision offer promising solutions to improve efficiency and scalability~\citep{fahlgren2015lights,araus2014field}. These innovations present opportunities to revolutionize plant stress phenotyping, potentially leading to more objective, rapid, and large-scale assessments that could significantly boost agricultural productivity.

Plant stress phenotyping tasks primarily fall into three categories: identification, classification, and quantification. Our study adopts a comprehensive view of plant stress phenotyping, encompassing traditional stresses, pest infestations, and seed quality issues. Identification detects stress presence (e.g., drought, nutrient deficiency, pathogens, pests). Classification categorizes stress into expert-defined classes, while quantification measures stress severity or extent, including seed quality impacts. Each task requires sophisticated analytical methods~\citep{ubbens2017deep}.

Recent computer vision and machine learning advances offer new opportunities for automating plant stress phenotyping~\citep{pound2017deep}. However, developing specialized models for agricultural applications faces challenges, primarily due to the need for high-quality annotated data. Expert knowledge in plant pathology, entomology, and agronomy is required for annotation, making it costly and time-consuming, thus creating a bottleneck in model development.

Researchers have explored techniques to develop effective models with limited annotated data~\citep{ogidi2023benchmarking}, including transfer learning from large-scale datasets~\citep{mohanty2016using}, self-supervised learning on unlabeled plant images~\citep{ogidi2023benchmarking}, and leveraging vision foundation models~\citep{chen2023adapting}. Despite progress, challenges remain due to domain differences between general and agricultural imagery, and the need for specialized visual understanding in plant stress phenotyping. The fine-grained nature of agricultural tasks often requires more nuanced feature recognition than general pretrained models offer, highlighting the need for adaptive, domain-specific approaches.

With traditional specialized model development, obtaining sufficient annotated data becomes increasingly challenging for each new use case. Typically, thousands of annotated examples are required to build effective models for specific agricultural tasks. As shown in \tabref{tab:traditional_models}, researchers have employed various specialized techniques to address this issue, including transfer learning, hybrid models, and custom architectures tailored to specific agricultural tasks. While these specialized models achieve impressive performance scores (e.g., 94\% accuracy on Soybean Disease classification and 91\% F1 score on DeepWeeds weed identification), they require significant domain expertise and computational resources to implement effectively. The high performance of these traditional approaches demonstrates their effectiveness for specific tasks, though their development and deployment costs remain substantial challenges.

Recent advances in vision language models (VLMs) have shown promising results in image-text understanding tasks~\citep{alayrac2022flamingo, dai2023instructblip}. These models demonstrate remarkable capabilities in processing and understanding visual and textual information jointly~\citep{yu2023mm,bitton2023visit,wang2024muirbench}. Additionally, VLMs can perform few-shot in-context learning  (learning from a small number of examples without model updates), adapting to new tasks with just a few examples in the prompt~\citep{jiang2024many}. VLMs often require only a couple of examples to learn and perform well on new tasks, potentially offering a more efficient solution to the data scarcity problem in agricultural applications.

Our study explores VLMs' potential for plant stress phenotyping tasks, hypothesizing that their broad visual-textual understanding could benefit scenarios with limited annotated data. We present a comprehensive evaluation of state-of-the-art VLMs on identification, classification, and quantification of plant stresses, using a curated benchmark dataset reflecting real-world agricultural scenarios. As these models become increasingly accessible to agricultural stakeholders, our study assesses their reliability and effectiveness in specialized tasks, providing insights into their practical potential for plant stress phenotyping.

This study makes several key \textbf{contributions} to plant stress phenotyping using VLMs. First, it evaluates these models on plant stress phenotyping tasks, providing insights into their potential to overcome traditional approach limitations. Second, it introduces a curated benchmark dataset mirroring real-world agricultural scenarios. Third, the study analyzes few-shot in-context learning performance on specialized agricultural tasks. Fourth, it presents a comparative analysis of various state-of-the-art VLMs in plant stress phenotyping. Finally, the study provides a quantitative assessment of how example relevance impacts few-shot in-context learning, advancing our understanding of these models in agricultural applications.

\section{Methodology}
\label{sec:Methodology}
This section details our approach to evaluating vision language models (VLMs) for plant stress phenotyping tasks.

\subsection{Task Formulation}
We adopt the following taxonomy for plant stress phenotyping tasks, focusing on:

\noindent\textbf{Identification (I):} Determining the specific type of stress from predefined options (e.g., identifying bacterial blight in wheat, or identifying a specific weed).

\noindent\textbf{Classification (C):} Categorizing stress into distinct severity classes (e.g., classifying iron deficiency chlorosis in soybean leaves into low, medium, or high levels).

\noindent\textbf{Quantification (Q):} Measuring the extent or severity of stress numerically (e.g., percentage of leaf area affected by disease).

\subsection{AgEval Benchmark Dataset Curation}
\label{sec:datasets}
We curated a diverse dataset comprising 12 subsets, each targeting specific plant stress phenotyping tasks. This collection, compiled from open-source resources, covers three main categories: Identification, Classification, and Quantification. These categories encompass various aspects of plant stress, from seed quality to pest infestations, reflecting the diverse challenges in agricultural stress assessment. The details of these datasets are provided in \tabref{tab:dataset_classification}.
We sampled 100 images from each dataset, evenly distributed across classes. Further details provided in \figref{fig:tree_map}.

\noindent\textbf{Identification:}
We considered datasets addressing the identification of \textbf{seed morphology} variations due to stresses, \textbf{foliar diseases}, and \textbf{pests} including weeds and insects that cause stresses. The Durum Wheat Dataset~\citep{kaya2019towards} and Soybean Seeds dataset \citep{soyabeanseeds} support seed morphology tasks, which involve identifying stress-induced changes in seed characteristics. The Mango Leaf Disease Dataset~\citep{mangoleaf,ahmed2023mangoleafbd}, Bean Leaf Lesions Classification dataset \citep{beanleaflesions}, and Soybean Diseases dataset~\citep{ghosal2018explainable} enable foliar stress tasks, focusing on identifying diverse plant stresses affecting leaves, including diseases and adverse environmental conditions. The DeepWeeds dataset~\citep{olsen2019deepweeds} and Dangerous Farm Insects dataset \citep{dangerousinsects} facilitate pest identification tasks, which involve recognizing weeds and insects~\citep{feuer2024zero} that cause plant stress. These datasets and their associated tasks collectively contribute to assessing stress impacts on seed quality, disease management, and pest control strategies in agriculture.

\noindent\textbf{Classification:}
We considered datasets for classification of \textbf{disease severity} and \textbf{stress tolerance} into expert-defined classes. The YELLOW-RUST-19 dataset \citep{yellowrustwheat,hayit2021determination, hayit2023classification} and Fusarium Wilt Disease in Chickpea dataset \citep{fusariumchickpea, hayit2024knn, hayit2024severity} support disease severity tasks, classifying disease stages caused by pathogens based on color and shape changes. The Iron Deficiency Chlorosis (IDC) Soybean Dataset~\citep{naik2017real} enables stress tolerance tasks, classifying abiotic stress stages caused by factors like nutrient deficiency or drought. 

\noindent\textbf{Quantification:}
We considered datasets addressing quantification of \textbf{pest} populations and \textbf{disease} extent. The InsectCount dataset \cite{nieuwenhuizen2019raw} supports pest quantification tasks involving counting insects in field images to assess infestation levels and inform pest management decisions. The PlantDoc dataset~\citep{leafsegmentation,10.1145/3371158.3371196} enables disease quantification tasks, measuring plant stress by segmenting diseased areas in leaf images and quantifying the percentage of affected areas to assess severity and spread.

% #-----------------------------

\subsection{Model Selection and Evaluation}
We evaluated six vision language models (VLMs): three state-of-the-art models (GPT-4o~\citep{gpt4o}, Claude 3.5 Sonnet~\citep{sonnet35}, and Gemini 1.5 Pro~\citep{reid2024gemini}), two budget-friendly options (Claude 3 Haiku~\citep{haiku3} and Gemini 1.5 Flash~\citep{reid2024gemini}), and one open-source alternative (LLaVA v1.6 34B~\citep{liu2024visual}). This selection encompasses a range of commercially available and open-source options to provide a comprehensive evaluation.

We evaluate VLM performance using both zero-shot and few-shot approaches. Zero-shot testing reveals inherent model capabilities, while few-shot testing (with 1, 2, 4, and 8 examples) demonstrates the models' ability to adapt to new tasks with minimal examples. For few-shot evaluations, we randomly select examples from the dataset. It's important to note that our few-shot learning refers to in-context learning capability with a few examples, not fine-tuning in a few-shot manner~\citep{mosbach2023few}.

\subsection{Performance Metrics}
We evaluated model performance using task-specific metrics: F1-score for Identification (I) tasks, and Normalized Mean Absolute Error (NMAE) for both Classification (C) and Quantification (Q) tasks.
For identification tasks, we used the weighted F1-score to account for potential class imbalance:
{\small
\begin{equation}
\text{F1-score} = \frac{\sum_{i=1}^{c} 2 w_i \cdot \text{precision}_i \cdot \text{recall}i}{\sum_{i=1}^{c} w_i (\text{precision}_i + \text{recall}_i)}
\end{equation}}
where $c$ is the number of classes and $w_i$ is the weight of the $i$-th class, proportional to the number of samples in that class.
For classification and quantification tasks, we calculated NMAE as:
{\small
\begin{equation}
\text{NMAE} = \frac{\sum_{i=1}^n |y_i - \hat{y}_i|}{\max(y) - \min(y)} \cdot 100\%
\end{equation}}
where $n$ is the number of samples, $y_i$ is the true value, $\hat{y}_i$ is the predicted value, and $\max(y)$ and $\min(y)$ are the maximum and minimum labels in the dataset. Note that the labels for classification problem are mapped to their corresponding ordinal values to be able to calculating NMAE. 
To handle out-of-vocabulary predictions from vision language models in classification tasks, we assigned the worst possible score to unseen labels in the ordinal mapping. This approach ensures consistency in evaluating predictions across all models.

In addition to these task-specific metrics, we employ Mean Reciprocal Rank (MRR) to compare model performance across datasets. MRR is calculated separately for Identification (ID) and Classification/Quantification (CQ) datasets:
{\small
\begin{equation}
\text{MRR}k = \frac{1}{|M|} \sum_{j=1}^{|M|} \frac{1}{|D_k|} \sum_{i=1}^{|D_k|} \frac{1}{r_{k,i,j}}
\end{equation}}
where:
$k \in {\text{ID}, \text{CQ}}$;
$|M|$ is the number of models being compared;
$|D_k|$ is the number of datasets in category $k$; and
$r_{k,i,j}$ is the rank of model $j$ for the $i$-th dataset in category $k$.
Ranks are determined as:
\begin{equation}
r_{\text{ID},i,j} = \text{rank}(\text{F1-score}_{i,j}, \text{descending})
\end{equation}
\begin{equation}
r_{\text{CQ},i,j} = \text{rank}(\text{NMAE}_{i,j}, \text{ascending})
\end{equation}
MRR is a comparative metric that indicates relative performance among models. It is calculated separately for zero-shot ($s=0$) and 8-shot ($s=8$) settings:
{\small
\begin{equation}
\text{MRR}{k,s} = \frac{1}{|M|} \sum_{j=1}^{|M|} \frac{1}{|D_k|} \sum_{i=1}^{|D_k|} \frac{1}{r_{k,i,j,s}}
\end{equation}}
\subsubsection{Relevance of Examples in Few-Shot Learning}

This analysis explores the impact of example relevance in few-shot learning for identification tasks. We investigate how examples from the same category versus diverse examples affect vision language models' (VLMs) performance in predicting image labels. This study aims to understand whether examples from the same dataset (or in a real-world scenario, examples related to the farmer's actual input) positively influence predictions, and whether related information steers the model towards more accurate category-specific predictions. 

Our analysis utilizes data from previous experiment runs, where few-shot examples and their labels were logged for each input across different shot settings. 

Let $\mathcal{D} = \{(x_i, y_i)\}_{i=1}^n$ be the dataset, where $x_i$ is an input and $y_i$ its true label. For a k-shot setting, let $\mathcal{E}_k(x_i) = \{(e_j, l_j)\}_{j=1}^k$ be the set of examples provided in the prompt for input $x_i$, where $e_j$ is an example and $l_j$ its label, such that $e_j \neq x_i$ for all $j$.

\begin{definition}[Bullseye Shot]
A k-shot prompt for input $x_i$ is considered a bullseye shot if:
\begin{equation}
\exists j \in \{1, \ldots, k\} : l_j = y_i
\end{equation}
\end{definition}
Intuitively, a bullseye shot occurs when at least one of the k example images provided in the prompt belongs to the same class as the target image being evaluated.

\noindent Note: A bullseye shot requires at least one example to match the true label, not necessarily all k examples.

This definition helps us quantify the effectiveness of providing relevant examples, offering insights into how VLMs can rapidly adapt to specialized agricultural tasks with minimal context, and how related information might enhance prediction accuracy for specific categories.

For each dataset and shot number $k \in \{1, 2, 4, 8\}$, we partition $\mathcal{D}$ into bullseye ($\mathcal{B}_k$) and non-bullseye ($\mathcal{N}_k$) subsets:
{\small
\begin{equation}
\mathcal{B}_k = \{(x_i, y_i) \in \mathcal{D} : \exists j \in \{1, \ldots, k\}, l_j = y_i \text{ in } \mathcal{E}_k(x_i)\}
\end{equation}
\begin{equation}
\mathcal{N}_k = \mathcal{D} \setminus \mathcal{B}_k
\end{equation}
We evaluate the impact using the performance delta from 0-shot ($\Delta F1$):
\begin{equation}
\Delta F1_{\mathcal{B}_k} = F1(\mathcal{B}_k) - F1_0
\end{equation}
\begin{equation}
\Delta F1_{\mathcal{N}_k} = F1(\mathcal{N}_k) - F1_0
\end{equation}}
where $F1_0$ is the 0-shot F1-score and $F1(\cdot)$ is the F1-score on the respective subset. The average impact across all evaluated shot numbers is calculated as:
{\small
\begin{equation}
\overline{\Delta F1}_{\mathcal{B}} = \frac{1}{|\mathcal{K}|} \sum_{k \in \mathcal{K}} \Delta F1_{\mathcal{B}_k}
\end{equation}
\begin{equation}
\overline{\Delta F1}_{\mathcal{N}} = \frac{1}{|\mathcal{K}|} \sum_{k \in \mathcal{K}} \Delta F1_{\mathcal{N}_k}
\end{equation}
}where $\mathcal{K}$ is the set of all shot numbers evaluated.
This analysis quantifies the performance impact of having relevant examples in few-shot prompts.

\begin{table*}[t!]
\centering
\caption{0-shot Performance of VLMs on AgEval Benchmark, Models Sorted by Average Performance (Highest to Lowest)}
\label{tab:zero_shot_results}
\vspace{-1em}
\begin{subtable}{1.0\linewidth}
\centering
\caption{Identification - Metric: F1 Score (Higher is Better). 
\protect\colorbox{yellow!25}{\strut Highest} 
\protect\colorbox{gray!25}{\strut Second Highest}}
\label{tab:zero_shot_results_identification}
\footnotesize
\begin{tabular}{lccccccc}
\hline
Model & \multicolumn{2}{c}{Seed Morphology} & \multicolumn{3}{c}{Foliar Stress} & \multicolumn{2}{c}{Pests} \\
 & Durum Wheat & Soybean & Mango Leaf Disease & Bean Leaf Lesions & Soybean Diseases & Dangerous Insects & Weeds \\
\hline
Gemini-pro-1.5 & \cellcolor{yellow!25}55.56 & 26.24 & 42.91 & \cellcolor{yellow!25}77.22 & \cellcolor{yellow!25}21.78 & \cellcolor{gray!25}82.67 & \cellcolor{yellow!25}46.83 \\
GPT-4o & 55.1 & 19.0 & \cellcolor{yellow!25}58.41 & 65.92 & 3.7 & \cellcolor{yellow!25}82.79 & \cellcolor{gray!25}38.77 \\
Claude-3.5-sonnet & \cellcolor{yellow!25}55.56 & \cellcolor{yellow!25}38.7 & \cellcolor{gray!25}49.82 & 68.65 & 8.54 & 82.02 & 18.85 \\
Gemini-flash-1.5 & 53.64 & 24.58 & 42.85 & \cellcolor{gray!25}70.61 & \cellcolor{gray!25}14.41 & 80.38 & 32.83 \\
Claude-3-haiku & 36.06 & \cellcolor{gray!25}31.24 & 29.83 & 55.26 & 12.69 & 51.28 & 13.86 \\
LLaVA v1.6 34B & 40.56 & 13.74 & 13.63 & 44.03 & 8.54 & 18.54 & 8.68 \\
\hline
\end{tabular}
\end{subtable}
\smallskip
\begin{subtable}{1.0\linewidth}
\centering
\caption{Classification and Quantification - Metric: NMAE (Lower is Better). 
\protect\colorbox{yellow!25}{\strut Lowest} 
\protect\colorbox{gray!25}{\strut Second Lowest}}
\label{tab:zero_shot_results_classification_quantification}
\footnotesize
\begin{tabular}{lccccc}
\hline
Model & \multicolumn{2}{c}{Disease Severity} & Stress Tolerance  & Pest & Disease \\
 & Yellow Rust 19  & FUSARIUM 22 & IDC & InsectCount & PlantDoc \\
\hline
Claude-3.5-sonnet & \cellcolor{gray!25}22.29  & \cellcolor{yellow!25}18.25& 26.28 & \cellcolor{gray!25}16.25 & \cellcolor{gray!25}15.59 \\
GPT-4o & \cellcolor{yellow!25}17.19 &  37.0 &\cellcolor{yellow!25}18.88  & \cellcolor{yellow!25}15.8 & 18.14 \\
Gemini-flash-1.5 & 31.25  & \cellcolor{gray!25}24.0& \cellcolor{gray!25}19.39 & 16.32 & 21.22 \\
Gemini-pro-1.5 & 26.25 & 33.0 &30.87 &  29.0 & \cellcolor{yellow!25}9.57 \\
Claude-3-haiku & 37.08  & 25.75& 22.86 & 28.34 & 22.14 \\
LLaVA v1.6 34B & 35.94  & 30.6 & 25.51 & 26.19 & 41.72 \\
\hline
\end{tabular}
\end{subtable}
\end{table*}

\subsubsection{Intra-task Uniformity}
To evaluate the performance consistency of Vision Language Models (VLMs) across different classes within individual (identification) tasks, we employ the Coefficient of Variation (CV) of all classes' F1 scores. This analysis is crucial for understanding the robustness and reliability of VLMs in domain-specific applications, not just in agriculture but across various specialized fields. By quantifying performance variability, we can identify potential biases or gaps in model training that may impact real-world deployment.

For each model-dataset combination, we calculate the CV as follows:
{\small
\begin{equation}
CV = \frac{\sigma}{\mu} \cdot 100\%
\end{equation}
}where $\sigma$ is the standard deviation of F1 scores across classes, and $\mu$ is the mean F1 score. The CV provides a normalized measure of dispersion, allowing for comparison across datasets with different scales.

For each identification dataset $d$ and model $m$, we calculate:
{\small
\begin{equation}
CV_{d,m} = \frac{\sqrt{\frac{1}{n-1}\sum_{i=1}^n (F1_{i,d,m} - \overline{F1_{d,m}})^2}}{\overline{F1_{d,m}}} \cdot 100\%
\end{equation}
}where $n$ is the number of classes in dataset $d$, $F1_{i,d,m}$ is the F1 score for class $i$, and $\overline{F1_{d,m}}$ is the mean F1 score across all classes for dataset $d$ and model $m$.

We then compute average CVs across datasets and models:
{\small
\begin{equation}
\overline{CV_d} = \frac{1}{|M|}\sum_{m \in M} CV_{d,m}
\end{equation}
\begin{equation}
\overline{CV_m} = \frac{1}{|D|}\sum_{d \in D} CV_{d,m}
\end{equation}
}where $M$ is the set of all models and $D$ is the set of all datasets.

These metrics allow us to quantify and compare the consistency of model performance across classes, highlighting areas where VLMs excel and identifying potential challenges in specific classes or model behaviors. This analysis provides valuable insights into the adaptability and robustness of VLMs in handling diverse tasks, which is essential for their effective application in specialized domains beyond agriculture, such as medical imaging, industrial quality control, or environmental monitoring.

\subsection{Prompt Engineering}
\label{prompt_eng}
We designed task-specific prompts to guide the VLMs in performing the ICQ tasks. The prompts were structured to provide clear instructions and ensure consistent output formatting across all models.

\begin{Verbatim}[breaklines=true, breakanywhere=true, fontsize=\footnotesize]
Given the image, identify the class. Use the following list of possible classes for your prediction It should be one of the : {expected_classes}. Be attentive to subtle details as some classes may appear similar. Provide your answer in the following JSON format:
{"prediction": "class_name"}...
\end{Verbatim}

For identification and classification tasks, we used a universal prompt template shown above (unless otherwise stated). Specialized prompts used for quantification tasks are given in the Supplement. For few-shot scenarios, we prepend examples to these prompts, maintaining the same structure and format across all shot counts to ensured consistency in inputs given to multiple model. 

\begin{table*}[t!]
\centering
\caption{8-shot Performance of VLMs on AgEval Benchmark, Models Sorted by Average Performance (Highest to Lowest)}
\label{tab:eight_shot_results}
\vspace{-1em}
\begin{subtable}{1.0\linewidth}
\centering
\caption{Identification - Metric: F1 Score (Higher is Better). 
\protect\colorbox{yellow!25}{\strut Highest} 
\protect\colorbox{gray!25}{\strut Second Highest}}
\label{tab:eight_shot_results_identification}
\footnotesize
\begin{tabular}{lccccccc}
\hline
Model & \multicolumn{2}{c}{Seed Morphology} & \multicolumn{3}{c}{Foliar Stress} & \multicolumn{2}{c}{Pests} \\
 & Durum Wheat & Soybean & Mango Leaf Disease & Bean Leaf Lesions & Soybean Diseases & Dangerous Insects & Weeds \\
\hline
GPT-4o & \cellcolor{yellow!25}95.94 & 48.29 & \cellcolor{yellow!25}80.96 & \cellcolor{yellow!25}86.9 & \cellcolor{yellow!25}62.96 & 82.56 & \cellcolor{yellow!25}56.03 \\
Gemini-pro-1.5 & 79.66 & \cellcolor{yellow!25}52.19 & \cellcolor{gray!25}71.68 & 78.17 & \cellcolor{gray!25}24.41 & \cellcolor{yellow!25}82.98 & \cellcolor{gray!25}49.96 \\
Claude-3.5-sonnet & \cellcolor{gray!25}89.66 & \cellcolor{gray!25}51.17 & 61.68 & \cellcolor{gray!25}84.78 & 11.07 & 81.89 & 27.17 \\
Gemini-flash-1.5 & 83.7 & 48.09 & 64.66 & 73.42 & 23.67 & \cellcolor{gray!25}82.72 & 41.89 \\
Claude-3-haiku & 53.29 & 38.02 & 38.92 & 46.42 & 8.81 & 45.08 & 15.34 \\
LLaVA v1.6 34B & 46.8 & 23.1 & 22.84 & 48.5 & 10.53 & 12.08 & 13.23 \\
\hline
\end{tabular}
\end{subtable}
\smallskip
\begin{subtable}{1.0\linewidth}
\centering
\caption{Classification and Quantification - Metric: NMAE (Lower is Better). 
\protect\colorbox{yellow!25}{\strut Lowest} 
\protect\colorbox{gray!25}{\strut Second Lowest}}
\label{tab:eight_shot_results_classification_quantification}
\footnotesize
\begin{tabular}{lccccc}
\hline
Model & \multicolumn{2}{c}{Disease Severity} & Stress Tolerance  & Pest & Disease \\
 & Yellow Rust 19  & FUSARIUM 22 & IDC & InsectCount & PlantDoc \\
\hline
Claude-3.5-sonnet & \cellcolor{gray!25}16.04 & \cellcolor{yellow!25}14.0 & 16.84 & \cellcolor{yellow!25}5.75 & \cellcolor{gray!25}11.31 \\
Gemini-pro-1.5 & 17.08  & \cellcolor{gray!25}17.0& \cellcolor{yellow!25}12.04 & 9.57 & 13.04 \\
Gemini-flash-1.5 & 20.83  & 17.5& \cellcolor{gray!25}15.56 & \cellcolor{gray!25}6.11 & 12.92 \\
GPT-4o & \cellcolor{yellow!25}15.83  & 19.75 & 60.82& 6.84 & \cellcolor{yellow!25}10.93 \\
Claude-3-haiku & 25.69 & 21.75 & 23.06 & 19.16 & 17.57 \\
LLaVA v1.6 34B & 30.56  & 60.0& 60.82 & 13.18 & 26.28 \\
\hline
\end{tabular}
\end{subtable}
\end{table*}

\begin{table*}[t!]
\centering
\caption{Few Shot Learning: Impact of having at least 1 example with same category as ground truth (Bullseye example).\\
\protect\colorbox{green!25}{\strut Highest} 
\protect\colorbox{red!25}{\strut Lowest} across 1, 2, 4, and 8-shot settings for both Bullseye and Non-Bullseye. 
Average Impact values are in \textbf{bold}.}
\label{tab:Bullseye}
\footnotesize
\begin{tabular}{lccccccccccc}
\hline
 & Baseline & \multicolumn{5}{c}{Bullseye Shots} & \multicolumn{5}{c}{Non-Bullseye Shots} \\
 & 0-shot & 1-shot & 2-shot & 4-shot & 8-shot & Avg. Impact & 1-shot & 2-shot & 4-shot & 8-shot & Avg. Impact \\
\hline
Durum Wheat & 51.18 & \cellcolor{green!25}+31.10 & +23.74 & +28.02 & +30.67 & \textbf{+28.38} & -01.74 & +01.03 & -04.39 & \cellcolor{red!25}-11.85 & \textbf{-04.24} \\
Soybean Seeds & 27.95 & \cellcolor{green!25}+27.10 & +15.54 & +13.79 & +21.77 & \textbf{+19.55} & \cellcolor{red!25}-01.48 & +05.57 & +05.44 & +09.20 & \textbf{+04.68} \\
Mango Leaf Disease & 44.76 & +22.61 & +22.90 & +17.34 & \cellcolor{green!25}+28.30 & \textbf{+22.79} & -02.02 & \cellcolor{red!25}-02.96 & -00.40 & +03.31 & \textbf{-00.52} \\
Bean Leaf Lesions & 67.53 & \cellcolor{green!25}+11.46 & +05.01 & +07.67 & +06.90 & \textbf{+07.76} & -06.08 & -05.61 & \cellcolor{red!25}-09.65 & -07.87 & \textbf{-07.30} \\
Soybean Diseases & 12.23 & \cellcolor{green!25}+49.05 & +17.71 & +09.26 & +06.54 & \textbf{+20.64} & \cellcolor{red!25}+00.26 & +06.98 & +04.95 & +11.73 & \textbf{+05.98} \\
Dangerous Insects & 75.83 & -12.05 & -10.92 & \cellcolor{red!25}-12.67 & -04.15 & \textbf{-09.95} & -00.92 & +02.04 & \cellcolor{green!25}+03.37 & +01.90 & \textbf{+01.60} \\
DeepWeeds & 30.23 & +22.37 & \cellcolor{green!25}+24.79 & +13.85 & +12.86 & \textbf{+18.47} & -02.03 & \cellcolor{red!25}-03.32 & -00.35 & +01.55 & \textbf{-01.04} \\
Average & 44.24 & \cellcolor{green!25}+21.66 & +14.11 & +11.04 & +14.70 & \textbf{+15.38} & \cellcolor{red!25}-02.00 & +00.53 & -00.15 & +01.14 & \textbf{-00.12} \\
\hline
\end{tabular}
\end{table*}

\section{Results}
\subsection{Zero-shot Performance}
\begin{figure}[t!]
    \centering
    \includegraphics[width=0.74\linewidth,trim={0.0in 0.0in 0.0in 0.2in},clip]{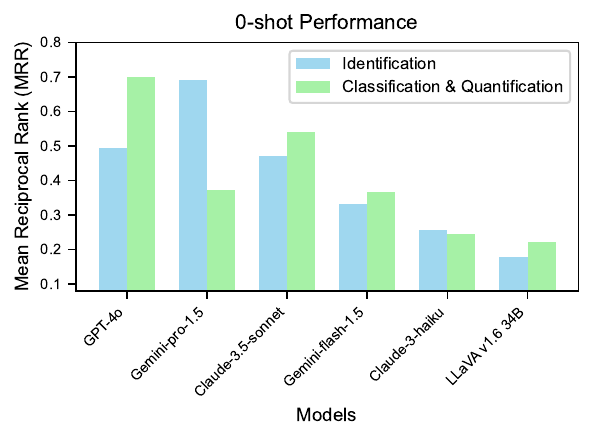}
    \caption{Zero-shot comparative performance of VLMs.}
    \label{fig:0_shot_performance}
\end{figure}

The zero-shot performance of Vision Language Models (VLMs) on the AgEval benchmark reveals interesting patterns (\figref{fig:0_shot_performance}). In identification tasks, Gemini-pro-1.5 demonstrates the strongest performance with an MRR of 0.69. For classification and quantification tasks, GPT-4o emerges as the top performer with an MRR of 0.70, closely followed by Claude-3.5-sonnet at 0.54. This highlights the varying strengths of different models across task types, emphasizing the importance of model selection based on specific requirements within the agricultural domain.

The strong performance of larger models like GPT-4o and Gemini-pro-1.5 suggests that general-purpose training translates well to domain-specific performance, underscoring the value of comprehensive pretraining in enabling VLMs to adapt to specialized agricultural tasks without additional fine-tuning.

\noindent\textbf{Identification Tasks (F1 Score):}
In zero-shot identification tasks, Gemini-pro-1.5 leads with an average F1 score of 50.45 across the 7 tasks, outperforming others in 4 out of 7 tasks. GPT-4o follows with 46.24 F1, showing strength in Mango Leaf Disease identification (58.41). Claude-3.5-sonnet performs competitively (46.02), particularly in Soybean Seeds identification (38.70).

\begin{figure}[t!]
    \centering
    \includegraphics[width=0.8\linewidth,trim={0.0in 0.0in 0.0in 0.2in},clip]{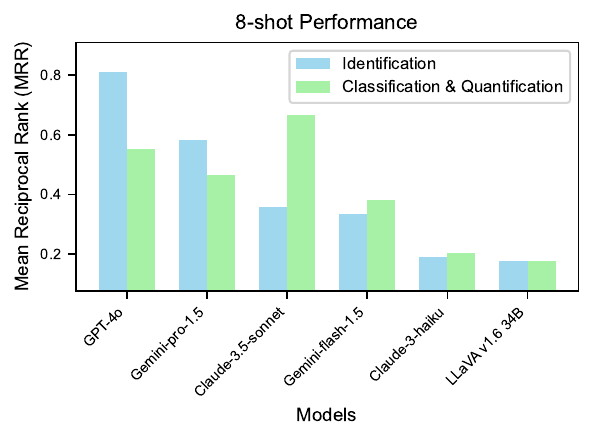}
    \caption{8-shot comparative performance of VLMs.}
    \label{fig:8_shot_performance}
\end{figure}

\noindent\textbf{Classification and Quantification Tasks (NMAE):}
For zero-shot classification and quantification, Claude-3.5-sonnet leads with the lowest average NMAE of 19.73 across 5 tasks. GPT-4o follows closely (21.40), outperforming in 3 out of 5 tasks. These results reinforce that larger VLMs can effectively leverage their general-purpose training to perform well on specialized agricultural tasks, even without domain-specific fine-tuning.

\begin{figure*}[t!]
    \centering
    \includegraphics[width=\textwidth]{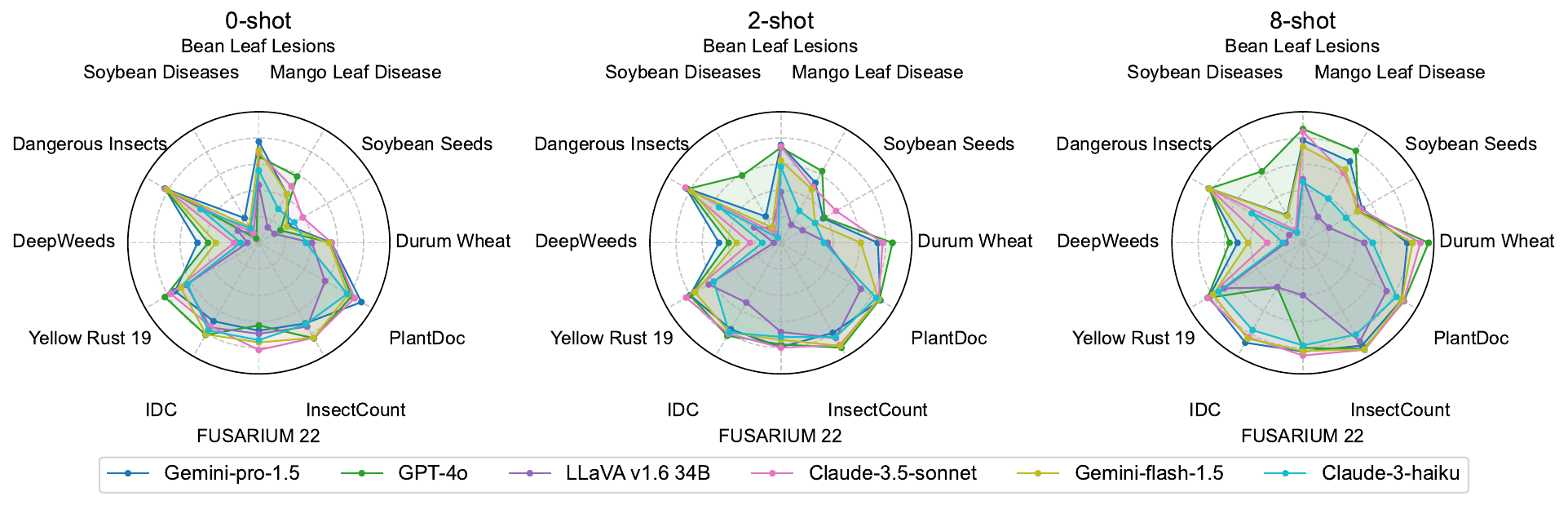}
    
    \caption{Performance comparison of models across 0-shot, 2-shot, and 8-shot settings on various datasets. F1 scores are shown directly, while NMAE is inverted (100 - NMAE) for consistent visualization, with higher values indicating better performance}
    \label{fig:radar_plots}
\end{figure*}

\begin{figure*}[t!]
    \centering
    \includegraphics[width=0.99\linewidth]{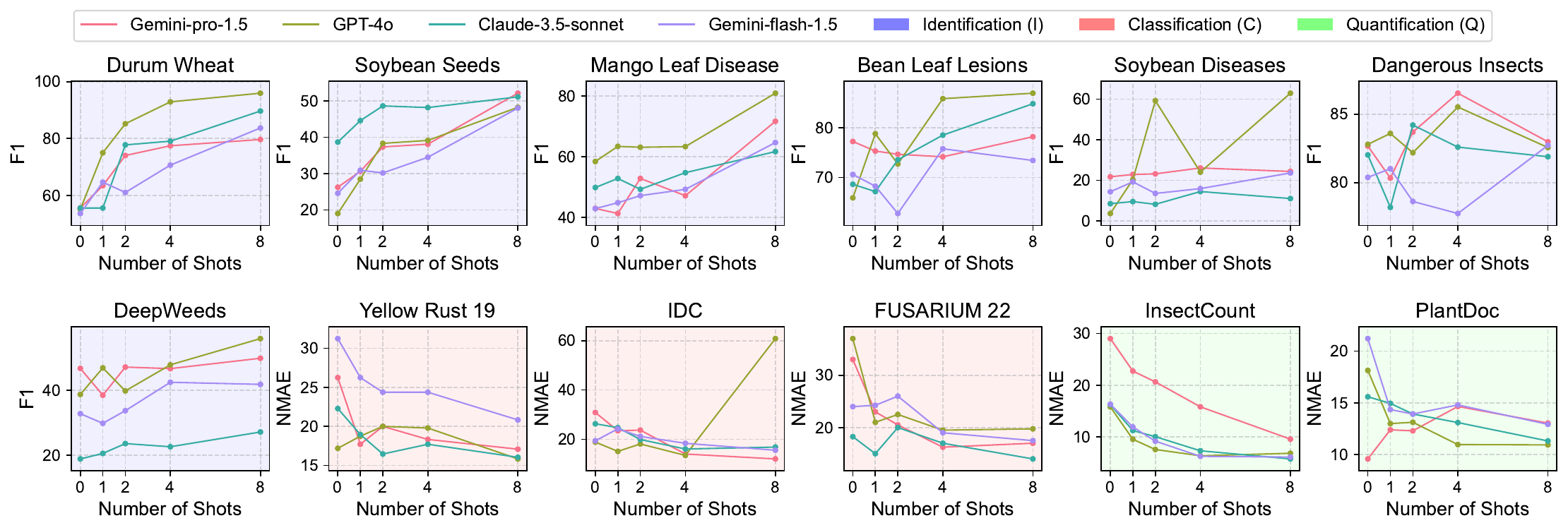}
    \caption{Performance comparison on individual tasks of the AgEval benchmark across different shot settings (0 to 8 shots) for top-4 performing VLMs.}
    \label{fig:individual-datasets}
\end{figure*}

\subsection{Full (8)-shot Performance}
The introduction of eight examples per task (8-shot learning) leads to significant improvements in model performance (\figref{fig:8_shot_performance}), achieving results that traditionally require thousands of annotated examples. For identification tasks, GPT-4o achieves the highest MRR of 0.81, a substantial increase from its zero-shot performance. In classification and quantification tasks, Claude-3.5-sonnet emerges as the top performer with an MRR of 0.66, while GPT-4o maintains strong performance with an MRR of 0.55.

Most models show improved performance with 8-shot learning compared to zero-shot, particularly in complex tasks, highlighting the efficiency of VLMs in learning from limited data.

\noindent\textbf{Identification Tasks (F1 Score):}
In 8-shot identification, GPT-4o achieves an average F1 score of 73.37, outperforming in 5 out of 7 tasks. It excels in Durum Wheat (95.94) and Bean Leaf Lesions (86.90) identification. Gemini-pro-1.5 maintains strong performance (62.72), leading in DeepWeeds identification (49.96).

\noindent\textbf{Classification and Quantification Tasks (NMAE):}
For 8-shot classification and quantification, Claude-3.5-sonnet leads with an average NMAE of 12.78, excelling in 3 out of 5 tasks. Gemini-pro-1.5 shows substantial improvement (13.74), particularly in IDC classification (12.04).

LLaVA's performance unexpectedly deteriorates with increased shot numbers, suggesting potential limitations in its in-context learning capabilities for AgEval tasks. These results underscore the varying impacts of few-shot learning across models and tasks, highlighting the importance of model selection based on specific agricultural task requirements.

\subsection{Relevance of Examples in Few-Shot Learning}
The relevance of examples in few-shot learning significantly influences model performance across various identification tasks. As shown in \tabref{tab:Bullseye}, the presence of exact category examples (Bullseye) consistently improves F1 scores, with an average increase of 15.38\% across all shot settings. This impact is most pronounced in the 1-shot scenario (+21.66\%). The absence of exact category examples (Non-Bullseye) has a minimal overall impact (-0.12\% on average), suggesting VLM robustness to less relevant examples, especially as the number of shots increases.

The benefits of relevant examples persist across different shot counts, albeit with diminishing returns. The average Bullseye impact decreases from +21.66\% in 1-shot to +14.70\% in 8-shot settings. Dataset-specific variations are substantial: Soybean Diseases shows the highest Bullseye impact (+49.05\% in 1-shot), while Bean Leaf Lesions exhibits more modest improvements (+11.46\% in 1-shot). The Dangerous Insects dataset presents an interesting case, showing slight negative impacts even with Bullseye examples (-9.95\% average impact), which may indicate domain-specific nuances. These findings underscore the importance of example selection in few-shot learning, particularly in low-shot scenarios. Please note that Llava was excluded from this analysis due to challenges in few-shot learning for our benchmark (see \figref{fig:individual-datasets}).

\subsection{Intra-task Uniformity}

\begin{figure}[b!]
    \centering
    \includegraphics[width=0.72\linewidth, trim={0.0in 0.3in 0.0in 0.3in},clip]{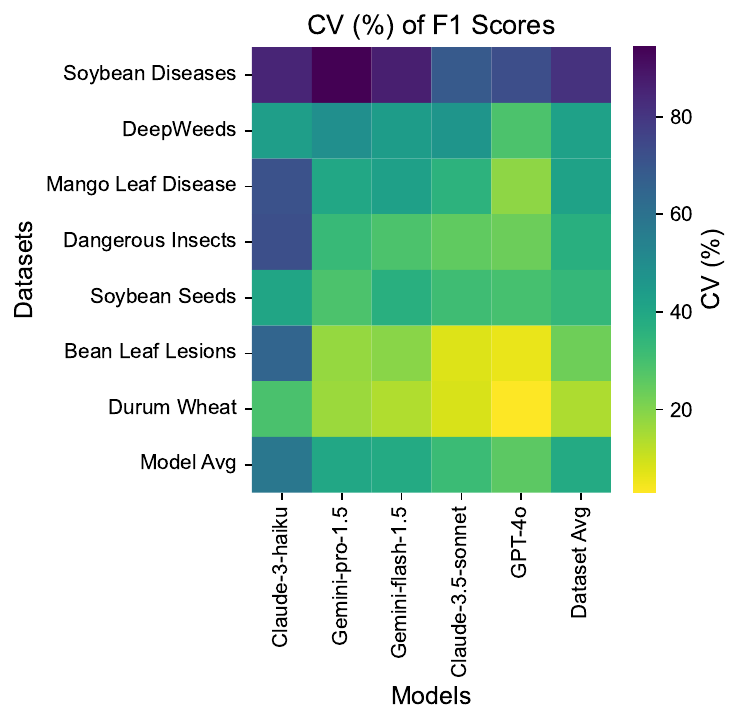}
    \caption{Heatmap of Coefficient of Variation (CV) for F1 scores (lower is better) across models and identification datasets.}
    \label{fig:cv_heatmap}
\end{figure}

Among VLMs, GPT-4o demonstrated the most consistent performance (\figref{fig:cv_heatmap}) with the lowest average CV (26\%), while Claude-3-haiku showed higher variability (CV=58\%). Gemini-pro-1.5 and Gemini-flash-1.5 exhibited moderate consistency (CV$\approx$39\%), with Claude-3.5-sonnet performing slightly better (CV=32\%). 

Regarding datasets, Soybean Diseases exhibited the highest average CV (81.29\%), indicating variability in model performance across its classes, potentially due to its low image resolution. Conversely, Durum Wheat showed the lowest average CV (14.28\%), implying more uniform performance. DeepWeeds and Mango Leaf Disease also demonstrated higher variability (CV \textgreater 40\%), highlighting areas for targeted improvements in VLM training to enhance performance uniformity.

These findings underscore the importance of considering not just overall accuracy, but also consistency across classes when selecting models for agricultural identification tasks. Detailed plots for each dataset's classes are provided in the supplement.

\subsection{Key Findings}
Here are 4 key findings from our evaluation: 

\noindent (1) GPT-4o demonstrates strong adaptability across AgEval tasks, showing significant improvement with few-shot learning (F1 score increase from 46.24 to 73.37). 

\noindent (2) Example relevance significantly impacts few-shot learning: On average, exact category examples (bullseyes) improve F1 scores by 15.38\%, while related examples from different classes have minimal impact. 

\noindent (3) VLM performance within datasets shows some variation: Coefficient of Variation ranges from 26.02\% (GPT-4o) to 58.03\% (Claude-3-haiku), indicating opportunities for further refinement in achieving consistent accuracy across all classes in plant stress phenotyping tasks. 

\noindent (4) Different models exhibit complementary strengths: Gemini-pro-1.5 excels in zero-shot identification (MRR 0.69), while GPT-4o leads in zero-shot classification/quantification (MRR 0.70). Claude Sonnet-3.5 consistently performs well in classification/quantification tasks, showcasing the diverse capabilities of VLMs in agricultural applications.

%------------------------------------------------------------------------
\section{Conclusions}
This study introduces AgEval, a benchmark for evaluating Vision Language Models (VLMs) on plant stress phenotyping tasks. We assembled diverse tasks across crops and stress types. Our evaluation examines zero-shot and few-shot performance, example relevance impact, and performance consistency using Coefficient of Variation. Results show VLMs' potential in addressing plant stress phenotyping challenges, with complementary strengths across models and tasks. This work establishes a baseline for future VLMs in agricultural contexts.

VLMs demonstrate adaptability to specialized agricultural tasks, with improvements in few-shot learning. Intra-task uniformity variation highlights refinement opportunities. These findings show VLMs are scalable solutions for plant stress phenotyping with minimal context. While VLMs may not match specialized models' peak performance, they offer valuable flexibility and reduced data requirements for practical agricultural applications

Future research could expand to broader agricultural tasks beyond plant stress phenotyping. Exploring increased shot counts could provide further insights. Fine-tuning models could enhance non-bullseye example performance and improve intra-task uniformity. Assessing practical deployment aspects in real-world settings, including computational requirements, data update strategies, integration with existing systems, and environmental and privacy considerations, will be crucial.

%%%%%%%%% REFERENCES

\clearpage 
{\small
\bibliographystyle{unsrtnat}
\bibliography{references}
}

\renewcommand{\thepage}{S\arabic{page}}
\renewcommand{\thesection}{S\arabic{section}}
\renewcommand{\thetable}{S\arabic{table}}
\renewcommand{\thefigure}{S\arabic{figure}}
\setcounter{figure}{0}
\setcounter{table}{0}
\setcounter{page}{0}
\setcounter{section}{0}
\clearpage

\section{Supplementary Material}
This section provides additional details about the datasets used in this study, including their names, links, and the classes they contain.

% Continue this pattern for the remaining datasets

\subsection{Prompts}
\label{supp:prompts}

For identification tasks, we used a universal prompt template, which was provided in the prompt engineering section, asking models to identify the class from a given list and provide the answer in JSON format. For classification and quantification tasks, we employed specialized prompts tailored to each dataset's requirements. These prompts included specific instructions on rating scales or counting methods relevant to the task at hand.

\noindent \textit{IDC Dataset:}

\begin{Verbatim}[breaklines=true, breakanywhere=true, fontsize=\footnotesize]
Analyze this image of a soybean canopy to determine the iron deficiency chlorosis (IDC) severity rating. The images are of soybean plants exhibiting various levels of IDC symptoms, ranging from healthy green plants to those with severe chlorosis and necrosis. Evaluate the extent of yellowing and browning in the canopy. Provide your answer in the following JSON format:
{{"prediction": "number"}}
Replace "number" with your best estimate of the IDC severity rating based on your analysis of the image.
The number should be entered exactly as a whole number (without any symbols) in a range of {expected_classes}. Higher value means more severity.
The response should start with {{ and contain only a JSON object (as specified above) and no other text.
\end{Verbatim}

\noindent \textit{Insect Count:}
\begin{Verbatim}[breaklines=true, breakanywhere=true, fontsize=\footnotesize]
Analyze this image of a yellow sticky insect trap. Count the total number of visible insects caught on the trap. Only look for insects which are easily visible to naked eye and look bigger compared to the other background artifacts. Provide your answer in the following JSON format:
{"prediction": "number"}
Replace "number" with your best estimate of the total insect count based on your analysis of the image. The number should be entered exactly as a whole number (without any symbols) in a range of {expected_classes} The response should start with { and contain only a JSON object (as specified above) and no other text.
\end{Verbatim}

\pagebreak

\noindent PlantDoc \textit{(Disease Quantification)}
\begin{Verbatim}[breaklines=true, breakanywhere=true, fontsize=\footnotesize]
Analyze this image of a leaf to get the total percentage of affected leaf. The images are of several plant leaf-like Apple Scab Leaf, Apple rust leaf, Bell_pepper leaf spot, Corn leaf blight, Potato leaf early blight, etc. The affected area is: diseased leaf area / total image area. Provide your answer in the following JSON format:
{"prediction": "number"}
Replace "number" with your best estimate of the percent on your analysis of the image. The number should be entered exactly as a whole number (without any symbols) in a range of {expected_classes} The response should start with { and contain only a JSON object (as specified above) and no other text.
only a JSON object (as specified above) and no other text.
\end{Verbatim}

\subsection{Additional dataset details}
\tabref{tab:dataset_classification} provides a comprehensive overview of the datasets used in the AgEval benchmark. It categorizes each dataset based on its primary task (Identification, Classification, or Quantification) and subcategory (e.g., Seed Morphology, Foliar Stress, Pests). The table includes key information such as the number of images, classes, year of creation, geographical location, and the evaluation metric used. This diverse collection of datasets covers various aspects of plant stress phenotyping, ranging from seed quality assessment to disease severity classification across different crops and regions. \tabref{tab:traditional_models} provides a comparison of the performance of traditional models on these datasets.

\begin{table*}[h!]
\centering
\caption{Classification of Agricultural Image Datasets. Categories: I (Identification), C (Classification), Q (Quantification)}
\label{tab:dataset_classification}
\resizebox{\textwidth}{!}{
\begin{tabular}{l l p{2cm} p{4cm} l l l l}
\hline
\textbf{Dataset} & \textbf{Category} & \textbf{Subcategory} & \textbf{Description} & \textbf{\# of Classes} & \textbf{Year} & \textbf{Location} & \textbf{Metric} \\
\hline
Durum Wheat~\citep{durumwheat,kaya2019towards} & I & Seed Morphology & Wheat variety identification & 3 & 2019 & Turkey & F1 \\
Soybean Seeds \citep{soyabeanseeds}  & I & Seed Morphology & Soybean quality prediction & 5 & N/A & N/A &  F1 \\
Mango Leaf Disease~\citep{mangoleaf,ahmed2023mangoleafbd} & I & Foliar Stress & Mango leaf disease classification & 8 & 2022 & Bangladesh &  F1 \\
Bean Leaf Lesions~\citep{beanleaflesions} & I & Foliar Stress & Bean leaf lesion type classification & 3 & N/A & N/A & F1 \\
Soybean Diseases~\citep{ghosal2018explainable} & I & Foliar Stress & Soybean stress identification  & 9 & 2016 & United States & F1 \\
Dangerous Insects \cite{dangerousinsects} & I & Pests & Harmful insects identification & 15 & N/A & N/A & F1 \\
DeepWeeds \citep{deepweeds,olsen2019deepweeds} & I & Pests & Weeds species identification & 9 & 2019 & Australia & F1 \\
Yellow Rust 19 \citep{yellowrustwheat,hayit2021determination, hayit2023classification} & C & Disease Severity  & Wheat yellow rust severity & 6 & 2021 & Turkey & NMAE \\
FUSARIUM 22 \citep{fusariumchickpea,hayit2024knn, hayit2024severity} & C & Disease Severity & Chickpea fusarium wilt severity & 5 & 2023 &  Turkey & NMAE \\
IDC~\citep{naik2017real}  & C & Stress Tolerance & Soybean stress severity & 5 & 2015 & United States & NMAE \\
InsectCount \cite{nieuwenhuizen2019raw} & Q & Pest Count & Insect count in images & - & 2021-2022 & N/A & NMAE \\
PlantDoc \citep{leafsegmentation,10.1145/3371158.3371196}  & Q & Disease & Percentage of the leaf that is diseased & - & N/A & N/A & NMAE \\

\hline
\end{tabular}
}
\end{table*}

\begin{table*}[h!]
    \centering
    \caption{Performance of the Traditional Models on Agricultural Image Datasets}
    \label{tab:traditional_models}
    \footnotesize
    \begin{tabular}{llllll}
    \hline
    \textbf{Dataset} & \textbf{Method/Approach Used} & \textbf{Reported Metric (Score)} & \textbf{Train} & \textbf{Validation} & \textbf{Test} \\
    \hline
    Durum Wheat~\citep{durumwheat,kaya2019towards} & Transfer learning with EfficientNetB3 & F1 Score (100) & 227(70\%) & 49 (15\%) & 49 (15\%) \\
    Soybean Seeds \citep{soyabeanseeds} & Transfer learning with ResNet50 & Accuracy (89) & 4410(80\%) & - & 1103(20\%) \\
    Mango Leaf Disease~\citep{mangoleaf,ahmed2023mangoleafbd} & Transfer learning with EfficientNetB3 & Accuracy (100) & 3200(80\%) & 480(12\%) & 320(8\%)\\
    Bean Leaf Lesions~\citep{beanleaflesions} & Hybrid Model (ViT, SVM) & F1 score (91) & 974 (84\%) & 133 (11\%) & 60 (5\%)\\
    Soybean Diseases~\citep{ghosal2018explainable} & Convolutional neural network & Accuracy (94) & 53266(81\%) & 5918(9\%) & 6576(10\%) \\
    Dangerous Insects \cite{dangerousinsects} & Transfer learning with Xception & Accuracy (77) & 1272 (80\%) & 287 (18\%) & 32 (2\%)\\

    DeepWeeds \citep{deepweeds,olsen2019deepweeds} & Transfer learning with ResNet26 & F1 score (91) & 11205(64\%) & 2801(16\%) & 3501(20\%) \\
    Yellow Rust 19 \citep{yellowrustwheat,hayit2021determination, hayit2023classification} & CNN-CGLCM with SVM & Accuracy (92) & 13500 (90\%) & - & 1500 (10\%) \\
    FUSARIUM 22 \citep{fusariumchickpea,hayit2024knn, hayit2024severity} & Hybrid Classifier (ViT,CatBoost) & F1 score (75) & 2950(68\%) & 521 (12\%) & 868(20\%) \\
    IDC~\citep{naik2017real} & Hierarchical classification & Accuracy (96) & 1479(75\%) & - & 493 (25\%) \\
    InsectCount \cite{nieuwenhuizen2019raw} & Internal dataset & \multicolumn{4}{c}{No baseline published} \\
    PlantDoc \citep{leafsegmentation,10.1145/3371158.3371196} & \multicolumn{5}{c}{No baseline exists on this data for this task} \\
    \hline
    \end{tabular}
\end{table*}

\begin{figure*}[t!]
    \centering
    \includegraphics[width=0.99\linewidth]{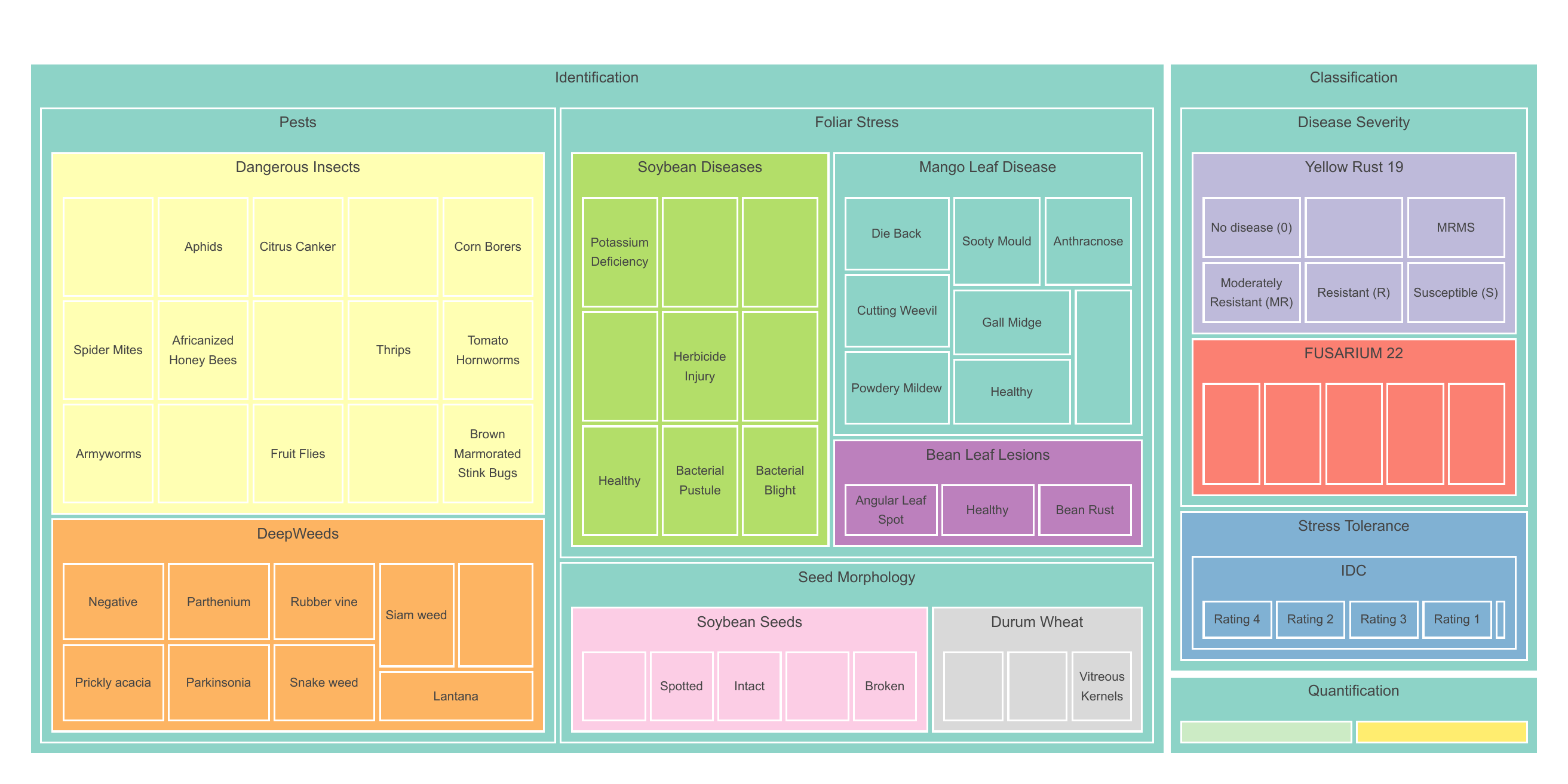}
    \caption{ Visualization of AgEval Benchmark Dataset - This treemap illustrates the distribution of datasets used in AgEval for plant stress identification, classification, and quantification. It contains subcategories, dataset names, and specific class names. Each rectangle represents a unique class name, with its size proportional to the count of instances. The visualization demonstrates the diversity of plant stress-related tasks covered by the AgEval framework across various crops and conditions.}
    \label{fig:tree_map}
\end{figure*}

\figref{fig:tree_map} provides a treemap visualization of the AgEval benchmark datasets, illustrating the distribution and hierarchy of tasks, subcategories, and individual classes. This comprehensive view highlights the diverse range of plant stress-related challenges addressed by AgEval, for all the AgEval benchmark. The size of each rectangle corresponds to the number of instances in that class, offering insights into the dataset composition and balance. We sampled 100 images in total from each dataset and the size corresponds to the resulting number of instances per class in each dataset used to build AgEval.

\subsection{Additional details on intra-task uniformity}
\begin{figure*}[t]
    \centering
    \begin{subfigure}[b]{0.49\linewidth}
        \centering
        \includegraphics[width=\linewidth]{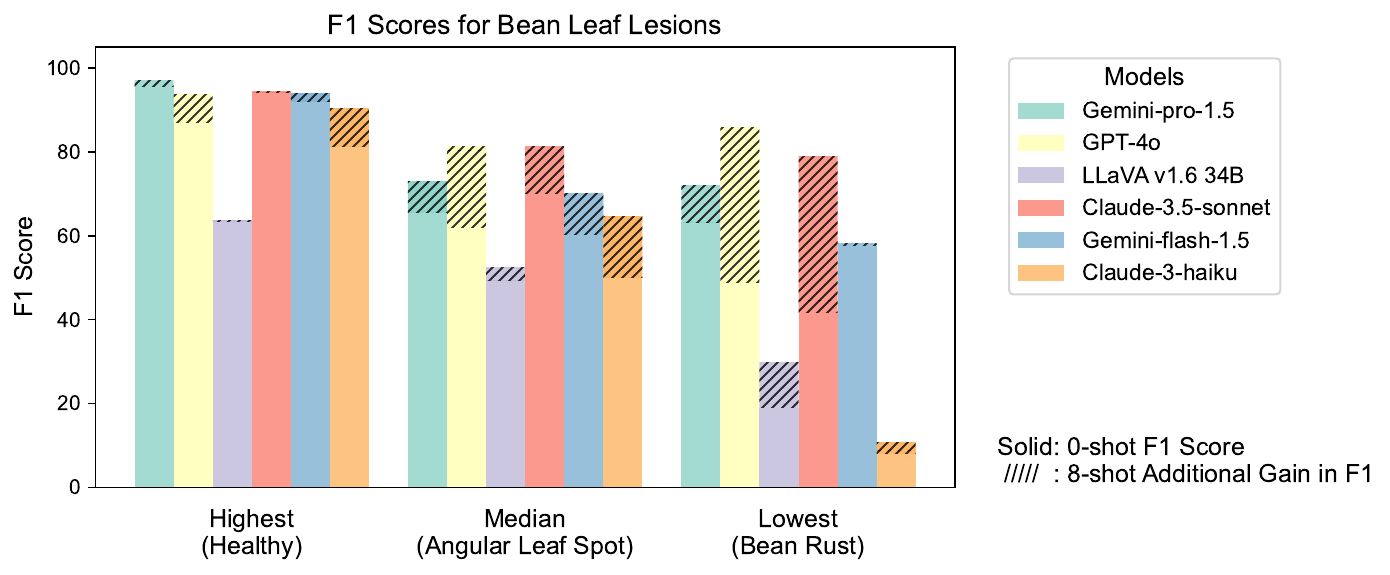}
        \caption{Bean Leaf Lesions F1 Scores}
        \label{fig:bean_leaf_lesions}
    \end{subfigure}
    \begin{subfigure}[b]{0.49\linewidth}
        \centering
        \includegraphics[width=\linewidth]{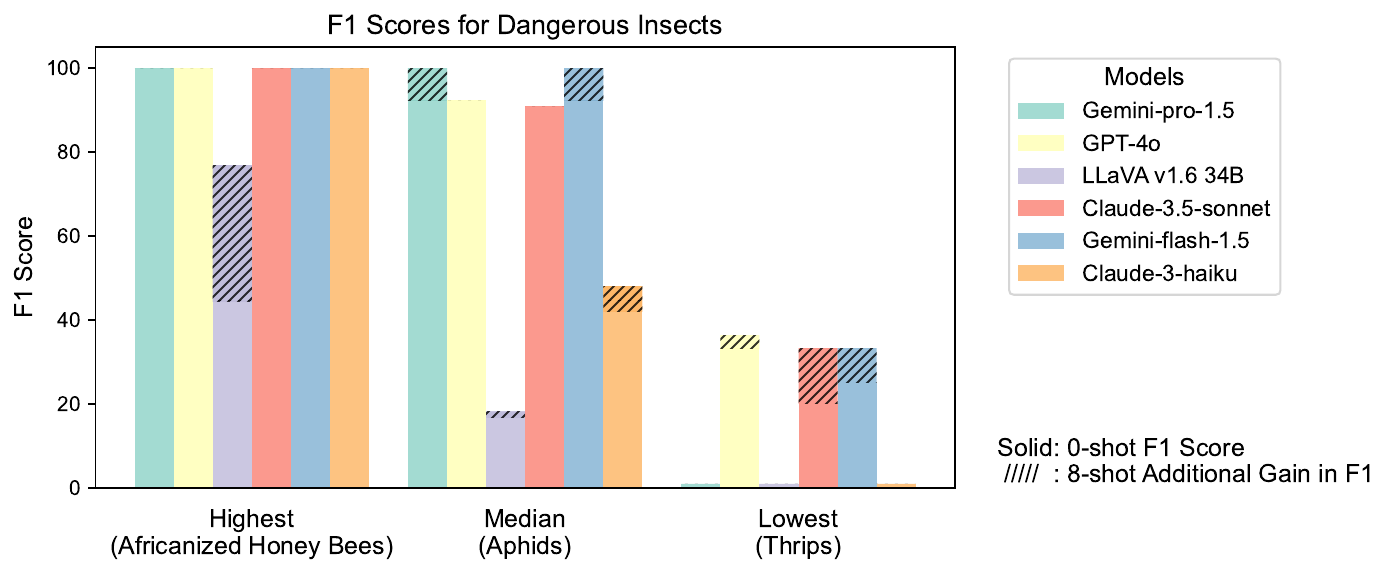}
        \caption{Dangerous Insects F1 Scores}
        \label{fig:dangerous_insects}
    \end{subfigure}
    \begin{subfigure}[b]{0.49\linewidth}
        \centering
        \includegraphics[width=\linewidth]{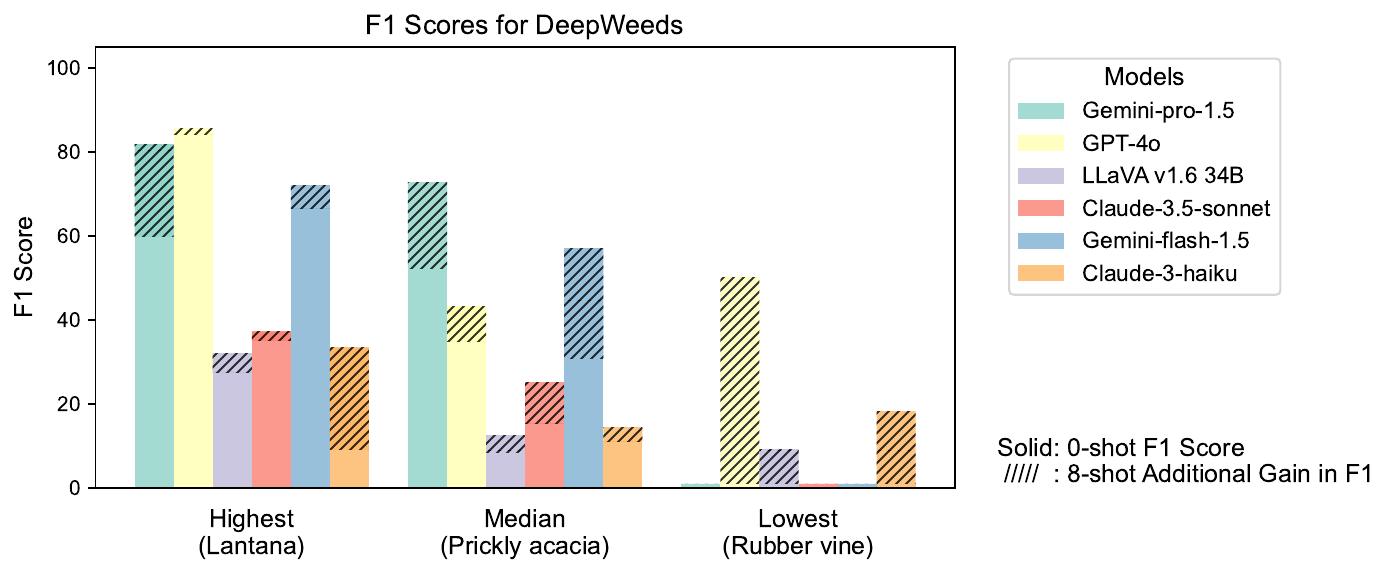}
        \caption{DeepWeeds F1 Scores}
        \label{fig:deepweeds}
    \end{subfigure}
    \label{fig:f1_scores_comparison1}
    \begin{subfigure}[b]{0.49\linewidth}
        \centering
        \includegraphics[width=\linewidth]{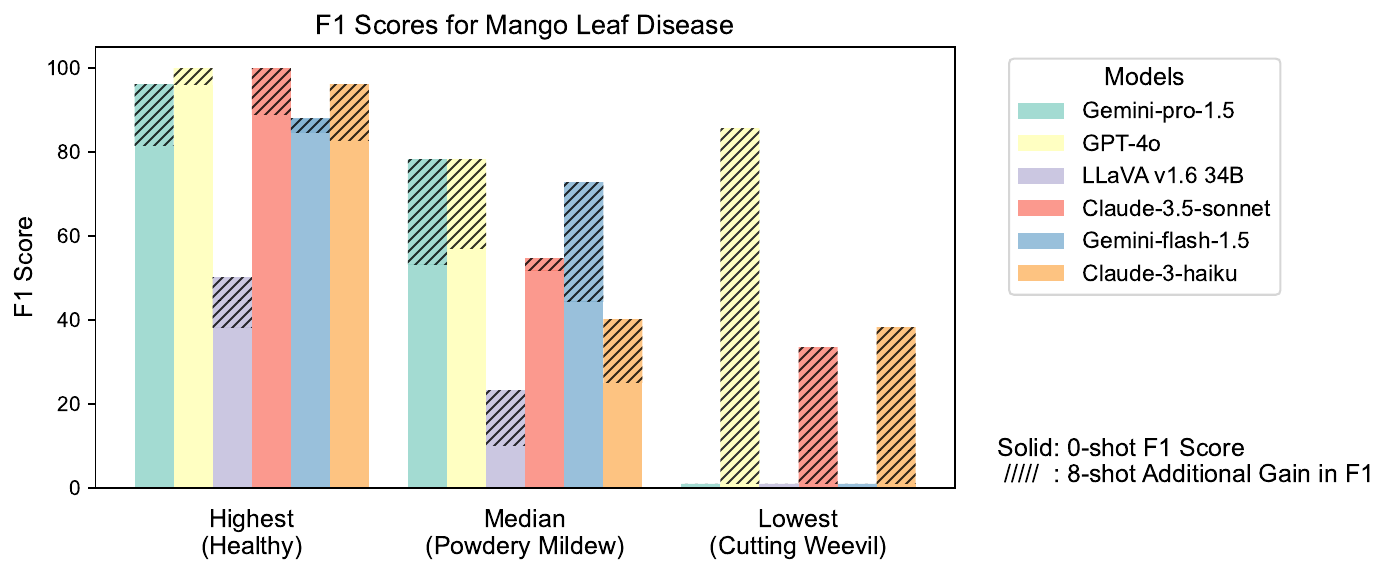}
        \caption{Mango Leaf Disease F1 Scores}
        \label{fig:mango_leaf_disease}
    \end{subfigure}
    \begin{subfigure}[b]{0.49\linewidth}
        \centering
        \includegraphics[width=\linewidth]{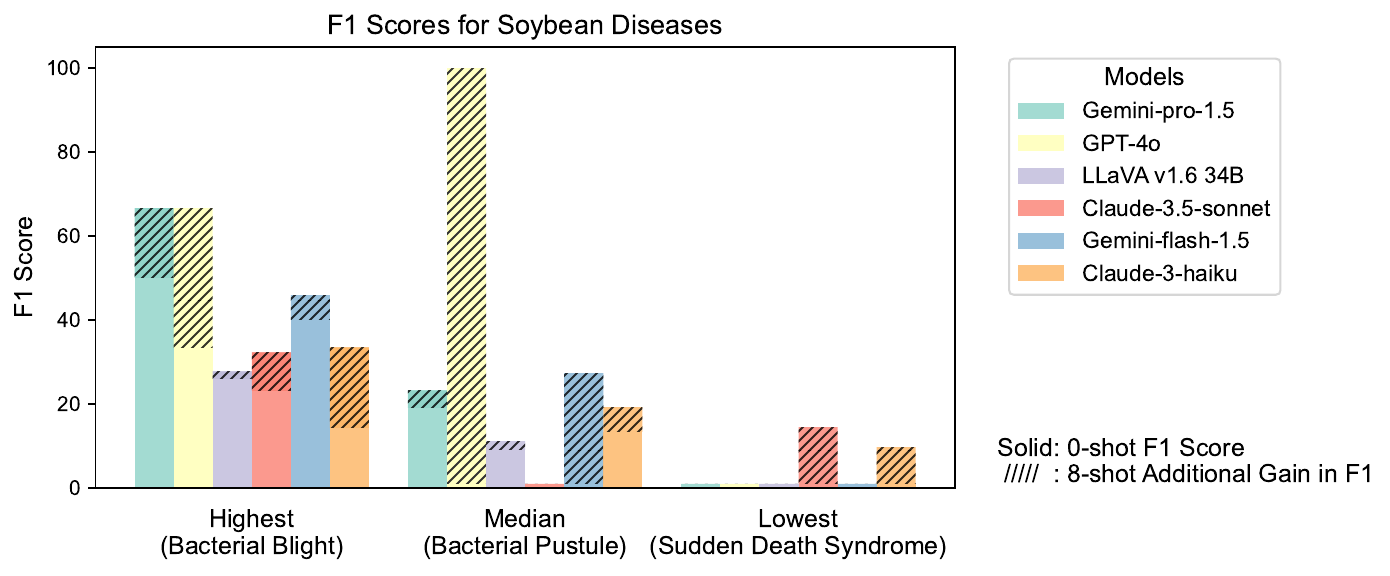}
        \caption{Soybean Diseases F1 Scores}
        \label{fig:soybean_diseases}
    \end{subfigure}
    \begin{subfigure}[b]{0.49\linewidth}
        \centering
        \includegraphics[width=\linewidth]{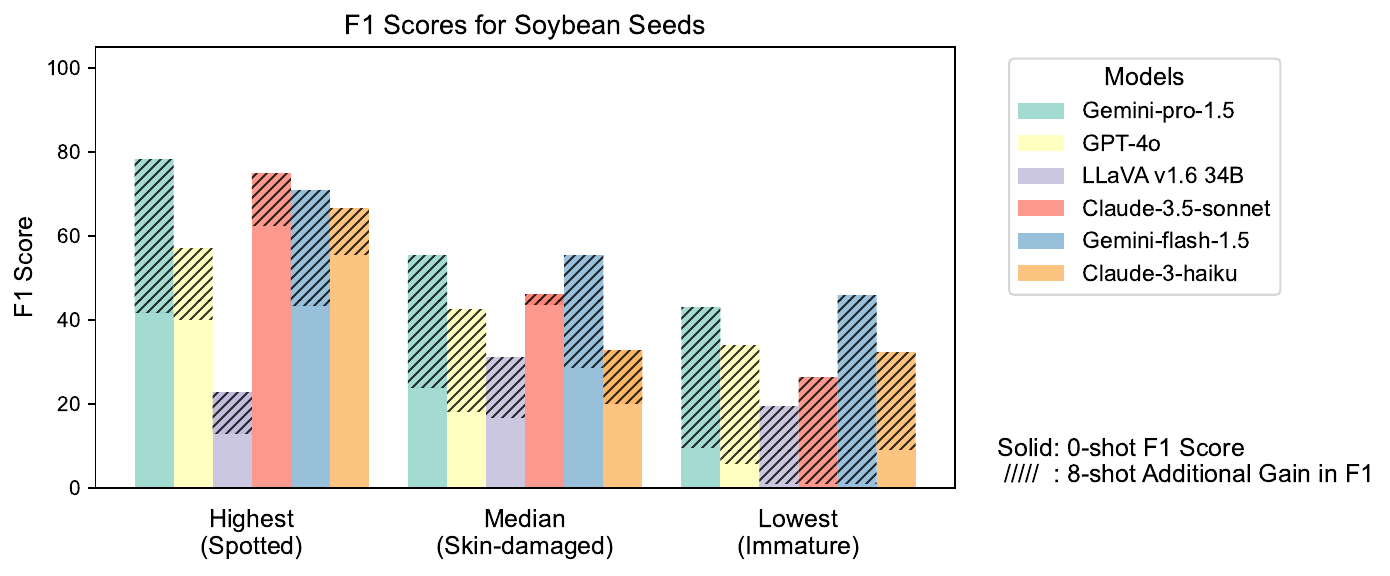}
        \caption{Soybean Seeds F1 Scores}
        \label{fig:soybean_seeds}
    \end{subfigure}
    \begin{subfigure}[b]{0.49\linewidth}
        \centering
        \includegraphics[width=\linewidth]{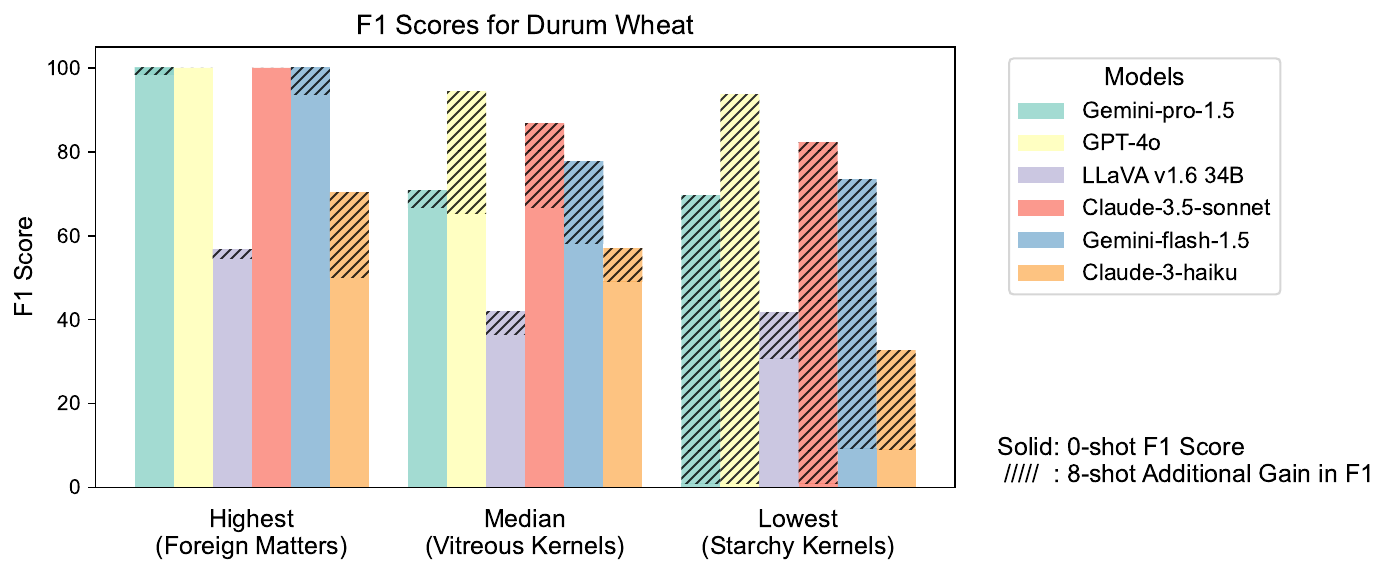}
        \caption{Durum Wheat F1 Scores}
        \label{fig:durum_wheat}
    \end{subfigure}
    \caption{Comparison of F1 Scores for classes within datasets for Highest, Medium and Lowest performing class.}
    \label{fig:intra-task-sup}
\end{figure*}

\figref{fig:intra-task-sup} provides a detailed examination of intra-task uniformity across different datasets in the AgEval benchmark. Each subfigure represents a specific dataset, showcasing the F1 scores for the highest, median, and lowest performing classes based on 0-shot performance. The visualization for each class displays both the 0-shot F1 score (solid bars) and the additional gain in F1 score achieved with 8-shot learning (hatched bars) for all six evaluated models.
This comprehensive view highlights the significant performance disparities among classes within each task, supporting our finding that the coefficient of variance (CV) ranges from 26.02\% to 58.03\% across models. The stark differences between the highest and lowest performing classes underscore the need for subject matter expertise to achieve reliable performance, especially for "difficult" classes.

\subsection{Anecdotal Samples from Each Task:}
Two samples and their corresponding predictions with respect to 0 and 8 shot are provided later. Please note that the questions are for illustration and actual prompts provided are in \secref{supp:prompts}  

            \begin{figure}[t!]
                \small
                \textbf{What wheat variety is this?}
                \vspace{1em}
                \begin{center}
                \includegraphics[height=0.45\linewidth]{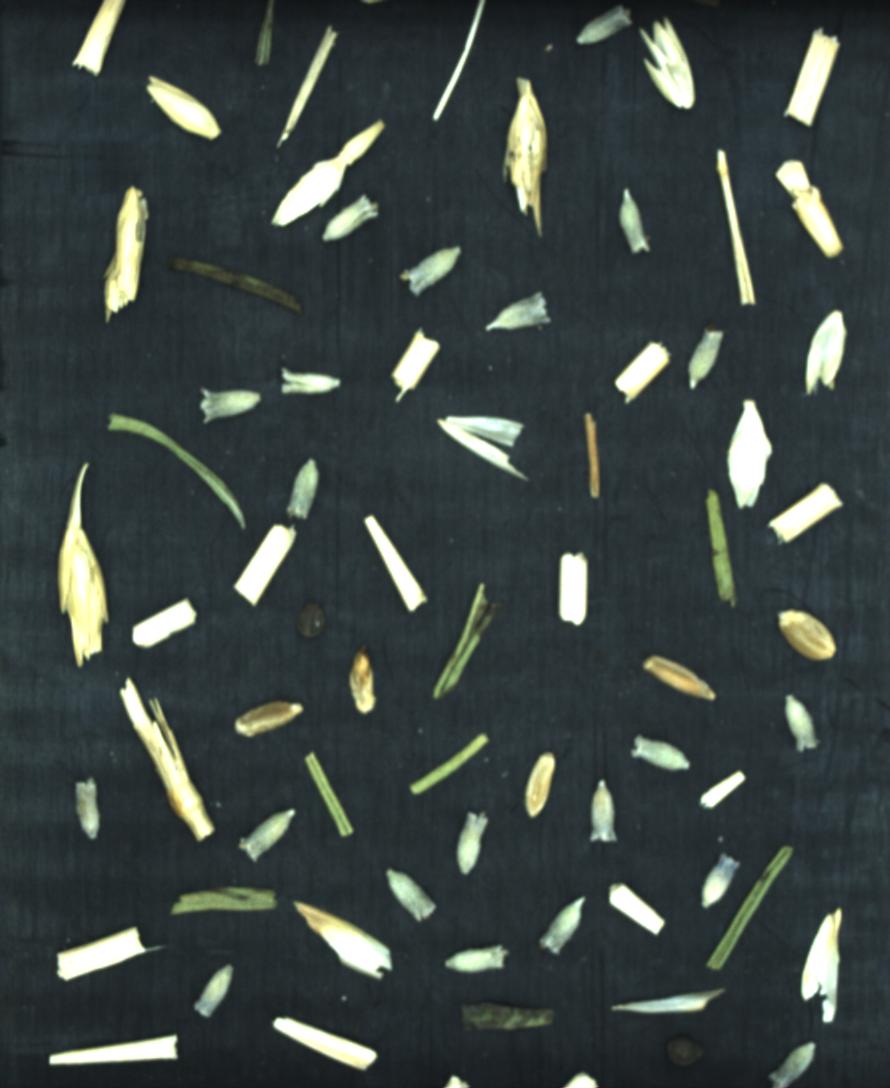}
                \end{center}
                \vspace{1em}
                
                \begin{tabular}{|p{0.28\linewidth}|p{0.28\linewidth}|p{0.28\linewidth}|}
                    \hline
                    \textbf{Category} & \textbf{Subcategory} & \textbf{Task} \\
                    \hline
                    Identification (I) & Seed Morphology & Durum Wheat \\
                    \hline
                \end{tabular}
                \vspace{1em}

                \textbf{Ground Truth:} Foreign Matters
                \vspace{1em}

                \textbf{Predictions:}
                \vspace{1em}
                
                \begin{tabular}{p{0.28\linewidth}|p{0.28\linewidth}|p{0.28\linewidth}}
                    \textbf{Model Name} & \textbf{0 shot} & \textbf{8 shot} \\
                    \hline
                    Gemini-pro-1.5 & Foreign Matters & Foreign Matters \\
                    GPT-4o & Foreign Matters & Foreign Matters \\
                    LLaVA v1.6 34B & Starchy Kernels & Vitreous Kernels \\
                    Claude-3.5-sonnet & Foreign Matters & Foreign Matters \\
                    Gemini-flash-1.5 & Foreign Matters & Foreign Matters \\
                    Claude-3-haiku & Vitreous Kernels & Vitreous Kernels
                \end{tabular}
            \end{figure}

            \begin{figure}[t!]
                \small
                \textbf{What wheat variety is this?}
                \vspace{1em}
                \begin{center}
                \includegraphics[height=0.45\linewidth]{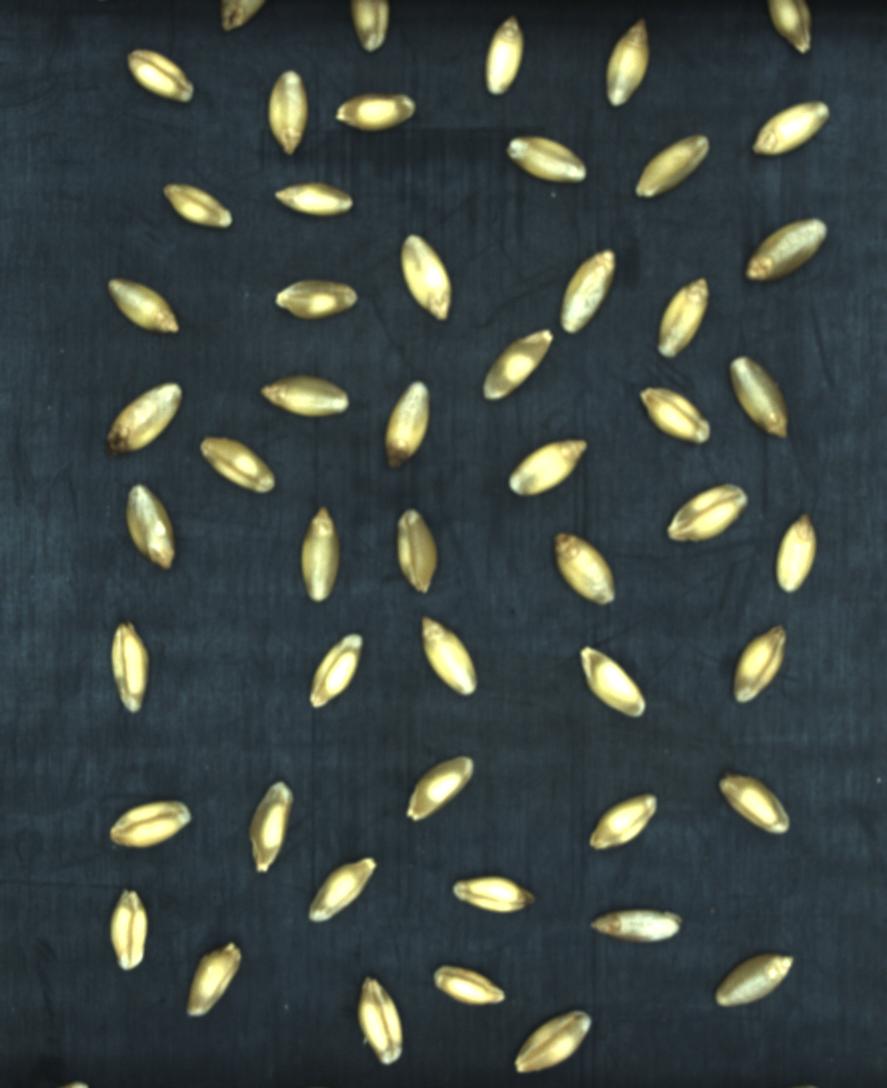}
                \end{center}
                \vspace{1em}
                
                \begin{tabular}{|p{0.28\linewidth}|p{0.28\linewidth}|p{0.28\linewidth}|}
                    \hline
                    \textbf{Category} & \textbf{Subcategory} & \textbf{Task} \\
                    \hline
                    Identification (I) & Seed Morphology & Durum Wheat \\
                    \hline
                \end{tabular}
                \vspace{1em}

                \textbf{Ground Truth:} Starchy Kernels
                \vspace{1em}

                \textbf{Predictions:}
                \vspace{1em}
                
                \begin{tabular}{p{0.28\linewidth}|p{0.28\linewidth}|p{0.28\linewidth}}
                    \textbf{Model Name} & \textbf{0 shot} & \textbf{8 shot} \\
                    \hline
                    Gemini-pro-1.5 & Vitreous Kernels & Starchy Kernels \\
                    GPT-4o & Vitreous Kernels & Starchy Kernels \\
                    LLaVA v1.6 34B & Vitreous Kernels & Vitreous Kernels \\
                    Claude-3.5-sonnet & Vitreous Kernels & Starchy Kernels \\
                    Gemini-flash-1.5 & Vitreous Kernels & Vitreous Kernels \\
                    Claude-3-haiku & Vitreous Kernels & Vitreous Kernels
                \end{tabular}
            \end{figure}

            \begin{figure}[t!]
                \small
                \textbf{What is the quality of the soybean seed?}
                \vspace{1em}
                \begin{center}
                \includegraphics[height=0.45\linewidth]{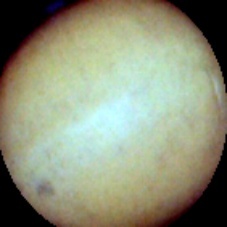}
                \end{center}
                \vspace{1em}
                
                \begin{tabular}{|p{0.28\linewidth}|p{0.28\linewidth}|p{0.28\linewidth}|}
                    \hline
                    \textbf{Category} & \textbf{Subcategory} & \textbf{Task} \\
                    \hline
                    Identification (I) & Seed Morphology & Soybean Seeds \\
                    \hline
                \end{tabular}
                \vspace{1em}

                \textbf{Ground Truth:} Intact
                \vspace{1em}

                \textbf{Predictions:}
                \vspace{1em}
                
                \begin{tabular}{p{0.28\linewidth}|p{0.28\linewidth}|p{0.28\linewidth}}
                    \textbf{Model Name} & \textbf{0 shot} & \textbf{8 shot} \\
                    \hline
                    Gemini-pro-1.5 & Spotted & Intact \\
                    GPT-4o & Spotted & Intact \\
                    LLaVA v1.6 34B & Immature & Intact \\
                    Claude-3.5-sonnet & Spotted & Intact \\
                    Gemini-flash-1.5 & Intact & Spotted \\
                    Claude-3-haiku & Intact & Intact
                \end{tabular}
            \end{figure}

            \begin{figure}[t!]
                \small
                \textbf{What is the quality of the soybean seed?}
                \vspace{1em}
                \begin{center}
                \includegraphics[height=0.45\linewidth]{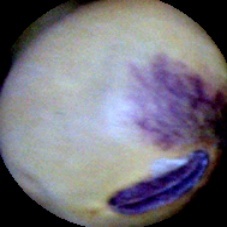}
                \end{center}
                \vspace{1em}
                
                \begin{tabular}{|p{0.28\linewidth}|p{0.28\linewidth}|p{0.28\linewidth}|}
                    \hline
                    \textbf{Category} & \textbf{Subcategory} & \textbf{Task} \\
                    \hline
                    Identification (I) & Seed Morphology & Soybean Seeds \\
                    \hline
                \end{tabular}
                \vspace{1em}

                \textbf{Ground Truth:} Spotted
                \vspace{1em}

                \textbf{Predictions:}
                \vspace{1em}
                
                \begin{tabular}{p{0.28\linewidth}|p{0.28\linewidth}|p{0.28\linewidth}}
                    \textbf{Model Name} & \textbf{0 shot} & \textbf{8 shot} \\
                    \hline
                    Gemini-pro-1.5 & Skin-damaged & Spotted \\
                    GPT-4o & Skin-damaged & Skin-damaged \\
                    LLaVA v1.6 34B & Skin-damaged & nan \\
                    Claude-3.5-sonnet & Spotted & Spotted \\
                    Gemini-flash-1.5 & Skin-damaged & Spotted \\
                    Claude-3-haiku & Skin-damaged & Spotted
                \end{tabular}
            \end{figure}

            \begin{figure}[t!]
                \small
                \textbf{What mango leaf disease is present?}
                \vspace{1em}
                \begin{center}
                \includegraphics[height=0.45\linewidth]{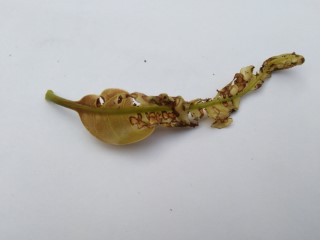}
                \end{center}
                \vspace{1em}
                
                \begin{tabular}{|p{0.28\linewidth}|p{0.28\linewidth}|p{0.28\linewidth}|}
                    \hline
                    \textbf{Category} & \textbf{Subcategory} & \textbf{Task} \\
                    \hline
                    Identification (I) & Foliar Stress & Mango Leaf Disease \\
                    \hline
                \end{tabular}
                \vspace{1em}

                \textbf{Ground Truth:} Anthracnose
                \vspace{1em}

                \textbf{Predictions:}
                \vspace{1em}
                
                \begin{tabular}{p{0.28\linewidth}|p{0.28\linewidth}|p{0.28\linewidth}}
                    \textbf{Model Name} & \textbf{0 shot} & \textbf{8 shot} \\
                    \hline
                    Gemini-pro-1.5 & Cutting Weevil & Bacterial Canker \\
                    GPT-4o & Cutting Weevil & Gall Midge \\
                    LLaVA v1.6 34B & Other & Anthracnose \\
                    Claude-3.5-sonnet & Cutting Weevil & Die Back \\
                    Gemini-flash-1.5 & nan & Anthracnose \\
                    Claude-3-haiku & Die Back & Anthracnose
                \end{tabular}
            \end{figure}

            \begin{figure}[t!]
                \small
                \textbf{What mango leaf disease is present?}
                \vspace{1em}
                \begin{center}
                \includegraphics[height=0.45\linewidth]{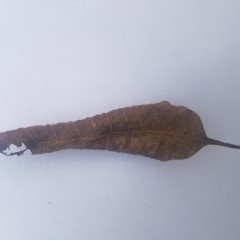}
                \end{center}
                \vspace{1em}
                
                \begin{tabular}{|p{0.28\linewidth}|p{0.28\linewidth}|p{0.28\linewidth}|}
                    \hline
                    \textbf{Category} & \textbf{Subcategory} & \textbf{Task} \\
                    \hline
                    Identification (I) & Foliar Stress & Mango Leaf Disease \\
                    \hline
                \end{tabular}
                \vspace{1em}

                \textbf{Ground Truth:} Die Back
                \vspace{1em}

                \textbf{Predictions:}
                \vspace{1em}
                
                \begin{tabular}{p{0.28\linewidth}|p{0.28\linewidth}|p{0.28\linewidth}}
                    \textbf{Model Name} & \textbf{0 shot} & \textbf{8 shot} \\
                    \hline
                    Gemini-pro-1.5 & nan & nan \\
                    GPT-4o & Die Back & Die Back \\
                    LLaVA v1.6 34B & Die Back & Bacterial Canker \\
                    Claude-3.5-sonnet & Die Back & Die Back \\
                    Gemini-flash-1.5 & nan & nan \\
                    Claude-3-haiku & Die Back & Die Back
                \end{tabular}
            \end{figure}

            \begin{figure}[t!]
                \small
                \textbf{What type of bean leaf lesion is this?}
                \vspace{1em}
                \begin{center}
                \includegraphics[height=0.45\linewidth]{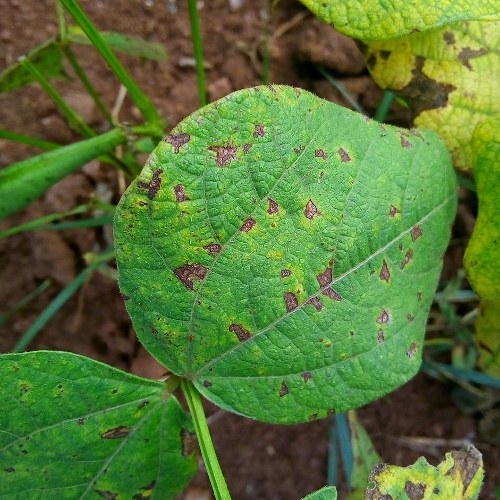}
                \end{center}
                \vspace{1em}
                
                \begin{tabular}{|p{0.28\linewidth}|p{0.28\linewidth}|p{0.28\linewidth}|}
                    \hline
                    \textbf{Category} & \textbf{Subcategory} & \textbf{Task} \\
                    \hline
                    Identification (I) & Foliar Stress & Bean Leaf Lesions \\
                    \hline
                \end{tabular}
                \vspace{1em}

                \textbf{Ground Truth:} Angular Leaf Spot
                \vspace{1em}

                \textbf{Predictions:}
                \vspace{1em}
                
                \begin{tabular}{p{0.28\linewidth}|p{0.28\linewidth}|p{0.28\linewidth}}
                    \textbf{Model Name} & \textbf{0 shot} & \textbf{8 shot} \\
                    \hline
                    Gemini-pro-1.5 & Angular Leaf Spot & Bean Rust \\
                    GPT-4o & Angular Leaf Spot & Angular Leaf Spot \\
                    LLaVA v1.6 34B & Bean Rust & Angular Leaf Spot \\
                    Claude-3.5-sonnet & Angular Leaf Spot & Angular Leaf Spot \\
                    Gemini-flash-1.5 & Angular Leaf Spot & Angular Leaf Spot \\
                    Claude-3-haiku & Angular Leaf Spot & Angular Leaf Spot
                \end{tabular}
            \end{figure}

            \begin{figure}[t!]
                \small
                \textbf{What type of bean leaf lesion is this?}
                \vspace{1em}
                \begin{center}
                \includegraphics[height=0.45\linewidth]{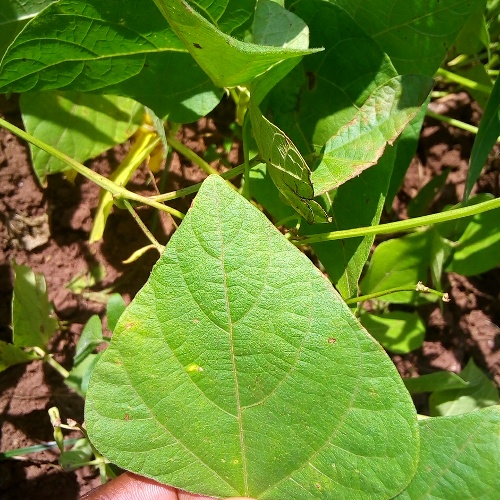}
                \end{center}
                \vspace{1em}
                
                \begin{tabular}{|p{0.28\linewidth}|p{0.28\linewidth}|p{0.28\linewidth}|}
                    \hline
                    \textbf{Category} & \textbf{Subcategory} & \textbf{Task} \\
                    \hline
                    Identification (I) & Foliar Stress & Bean Leaf Lesions \\
                    \hline
                \end{tabular}
                \vspace{1em}

                \textbf{Ground Truth:} Healthy
                \vspace{1em}

                \textbf{Predictions:}
                \vspace{1em}
                
                \begin{tabular}{p{0.28\linewidth}|p{0.28\linewidth}|p{0.28\linewidth}}
                    \textbf{Model Name} & \textbf{0 shot} & \textbf{8 shot} \\
                    \hline
                    Gemini-pro-1.5 & Healthy & Healthy \\
                    GPT-4o & Healthy & Healthy \\
                    LLaVA v1.6 34B & Healthy & Angular Leaf Spot \\
                    Claude-3.5-sonnet & Healthy & Bean Rust \\
                    Gemini-flash-1.5 & Healthy & Healthy \\
                    Claude-3-haiku & Healthy & Healthy
                \end{tabular}
            \end{figure}

            \begin{figure}[t!]
                \small
                \textbf{What is the type of stress in this soybean?}
                \vspace{1em}
                \begin{center}
                \includegraphics[height=0.4\linewidth]{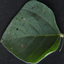}
                \end{center}
                \vspace{1em}
                
                \begin{tabular}{|p{0.28\linewidth}|p{0.28\linewidth}|p{0.28\linewidth}|}
                    \hline
                    \textbf{Category} & \textbf{Subcategory} & \textbf{Task} \\
                    \hline
                    Identification (I) & Foliar Stress & Soybean Diseases \\
                    \hline
                \end{tabular}
                \vspace{1em}

                \textbf{Ground Truth:} Frogeye Leaf Spot
                \vspace{1em}

                \textbf{Predictions:}
                \vspace{1em}
                
                \begin{tabular}{p{0.28\linewidth}|p{0.28\linewidth}|p{0.28\linewidth}}
                    \textbf{Model Name} & \textbf{0 shot} & \textbf{8 shot} \\
                    \hline
                    Gemini-pro-1.5 & Healthy & Potassium Deficiency \\
                    GPT-4o & Bacterial Pustule & Bacterial Blight \\
                    LLaVA v1.6 34B & Healthy & Iron Deficiency Chlorosis \\
                    Claude-3.5-sonnet & Healthy & Healthy \\
                    Gemini-flash-1.5 & Healthy & Healthy \\
                    Claude-3-haiku & Potassium Deficiency & Potassium Deficiency
                \end{tabular}
            \end{figure}

            \begin{figure}[t!]
                \small
                \textbf{What is the type of stress in this soybean?}
                \vspace{1em}
                \begin{center}
                \includegraphics[height=0.4\linewidth]{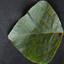}
                \end{center}
                \vspace{1em}
                
                \begin{tabular}{|p{0.28\linewidth}|p{0.28\linewidth}|p{0.28\linewidth}|}
                    \hline
                    \textbf{Category} & \textbf{Subcategory} & \textbf{Task} \\
                    \hline
                    Identification (I) & Foliar Stress & Soybean Diseases \\
                    \hline
                \end{tabular}
                \vspace{1em}

                \textbf{Ground Truth:} Bacterial Pustule
                \vspace{1em}

                \textbf{Predictions:}
                \vspace{1em}
                
                \begin{tabular}{p{0.28\linewidth}|p{0.28\linewidth}|p{0.28\linewidth}}
                    \textbf{Model Name} & \textbf{0 shot} & \textbf{8 shot} \\
                    \hline
                    Gemini-pro-1.5 & Herbicide Injury & Iron Deficiency Chlorosis \\
                    GPT-4o & Healthy & Bacterial Pustule \\
                    LLaVA v1.6 34B & Healthy & Healthy \\
                    Claude-3.5-sonnet & Healthy & Healthy \\
                    Gemini-flash-1.5 & Iron Deficiency Chlorosis & Healthy \\
                    Claude-3-haiku & Frogeye Leaf Spot & Sudden Death Syndrome
                \end{tabular}
            \end{figure}

            \begin{figure}[t!]
                \small
                \textbf{What is the name of this harmful insect?}
                \vspace{1em}
                \begin{center}
                \includegraphics[height=0.45\linewidth]{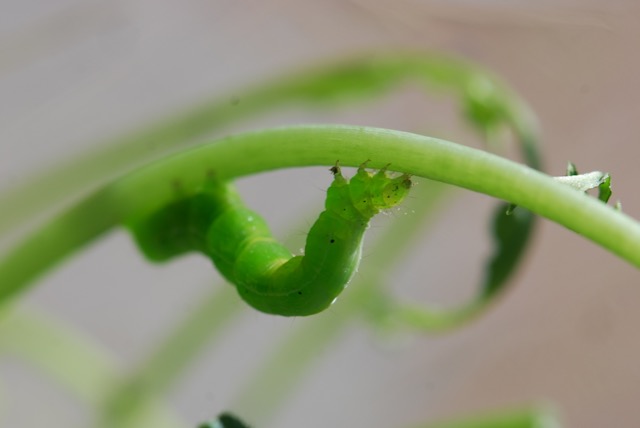}
                \end{center}
                \vspace{1em}
                
                \begin{tabular}{|p{0.28\linewidth}|p{0.28\linewidth}|p{0.28\linewidth}|}
                    \hline
                    \textbf{Category} & \textbf{Subcategory} & \textbf{Task} \\
                    \hline
                    Identification (I) & Invasive Species & Dangerous Insects \\
                    \hline
                \end{tabular}
                \vspace{1em}

                \textbf{Ground Truth:} Cabbage Loopers
                \vspace{1em}

                \textbf{Predictions:}
                \vspace{1em}
                
                \begin{tabular}{p{0.28\linewidth}|p{0.28\linewidth}|p{0.28\linewidth}}
                    \textbf{Model Name} & \textbf{0 shot} & \textbf{8 shot} \\
                    \hline
                    Gemini-pro-1.5 & Cabbage Loopers & Cabbage Loopers \\
                    GPT-4o & Cabbage Loopers & Cabbage Loopers \\
                    LLaVA v1.6 34B & Cabbage Loopers & nan \\
                    Claude-3.5-sonnet & Cabbage Loopers & Cabbage Loopers \\
                    Gemini-flash-1.5 & Cabbage Loopers & Cabbage Loopers \\
                    Claude-3-haiku & Aphids & Tomato Hornworms
                \end{tabular}
            \end{figure}

            \begin{figure}[t!]
                \small
                \textbf{What is the name of this harmful insect?}
                \vspace{1em}
                \begin{center}
                \includegraphics[height=0.45\linewidth]{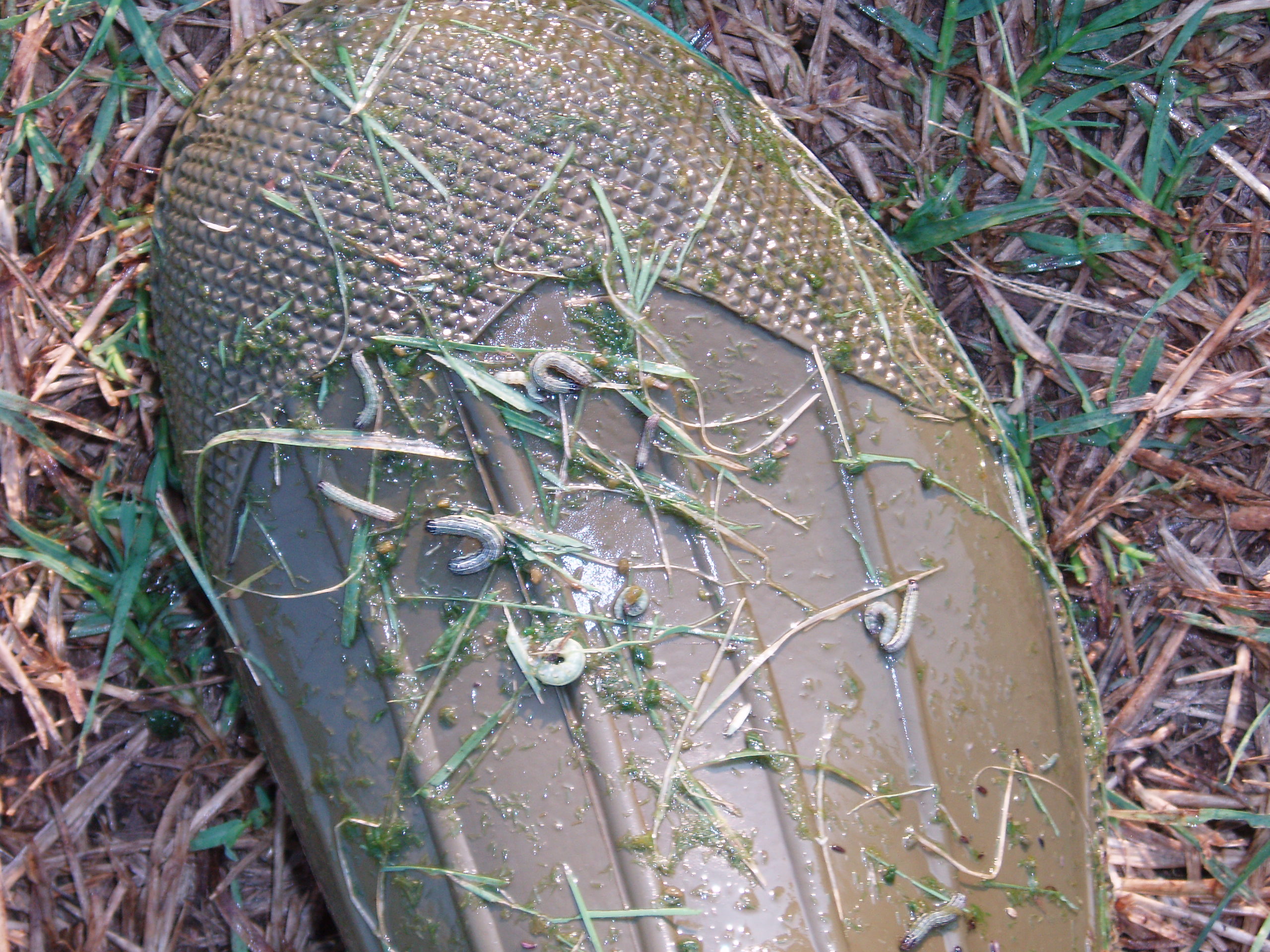}
                \end{center}
                \vspace{1em}
                
                \begin{tabular}{|p{0.28\linewidth}|p{0.28\linewidth}|p{0.28\linewidth}|}
                    \hline
                    \textbf{Category} & \textbf{Subcategory} & \textbf{Task} \\
                    \hline
                    Identification (I) & Invasive Species & Dangerous Insects \\
                    \hline
                \end{tabular}
                \vspace{1em}

                \textbf{Ground Truth:} Fall Armyworms
                \vspace{1em}

                \textbf{Predictions:}
                \vspace{1em}
                
                \begin{tabular}{p{0.28\linewidth}|p{0.28\linewidth}|p{0.28\linewidth}}
                    \textbf{Model Name} & \textbf{0 shot} & \textbf{8 shot} \\
                    \hline
                    Gemini-pro-1.5 & Armyworms & Armyworms \\
                    GPT-4o & Cabbage Loopers & Armyworms \\
                    LLaVA v1.6 34B & Cabbage Loopers & nan \\
                    Claude-3.5-sonnet & Armyworms & Armyworms \\
                    Gemini-flash-1.5 & Fall Armyworms & Armyworms \\
                    Claude-3-haiku & Armyworms & nan
                \end{tabular}
            \end{figure}

            \begin{figure}[t!]
                \small
                \textbf{What is the name of this weed?}
                \vspace{1em}
                \begin{center}
                \includegraphics[height=0.45\linewidth]{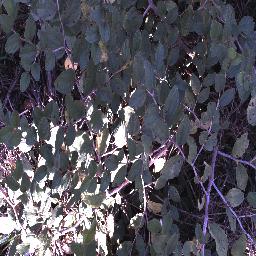}
                \end{center}
                \vspace{1em}
                
                \begin{tabular}{|p{0.28\linewidth}|p{0.28\linewidth}|p{0.28\linewidth}|}
                    \hline
                    \textbf{Category} & \textbf{Subcategory} & \textbf{Task} \\
                    \hline
                    Identification (I) & Invasive Species & DeepWeeds \\
                    \hline
                \end{tabular}
                \vspace{1em}

                \textbf{Ground Truth:} Chinee apple
                \vspace{1em}

                \textbf{Predictions:}
                \vspace{1em}
                
                \begin{tabular}{p{0.28\linewidth}|p{0.28\linewidth}|p{0.28\linewidth}}
                    \textbf{Model Name} & \textbf{0 shot} & \textbf{8 shot} \\
                    \hline
                    Gemini-pro-1.5 & Chinee apple & Chinee apple \\
                    GPT-4o & Chinee apple & Chinee apple \\
                    LLaVA v1.6 34B & Parthenium & Parkinsonia \\
                    Claude-3.5-sonnet & Lantana & Lantana \\
                    Gemini-flash-1.5 & Prickly acacia & Chinee apple \\
                    Claude-3-haiku & Parthenium & Parthenium
                \end{tabular}
            \end{figure}

            \begin{figure}[t!]
                \small
                \textbf{What is the name of this weed?}
                \vspace{1em}
                \begin{center}
                \includegraphics[height=0.45\linewidth]{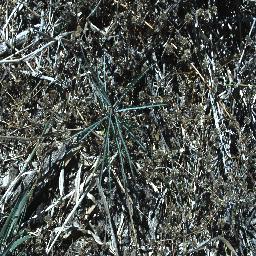}
                \end{center}
                \vspace{1em}
                
                \begin{tabular}{|p{0.28\linewidth}|p{0.28\linewidth}|p{0.28\linewidth}|}
                    \hline
                    \textbf{Category} & \textbf{Subcategory} & \textbf{Task} \\
                    \hline
                    Identification (I) & Invasive Species & DeepWeeds \\
                    \hline
                \end{tabular}
                \vspace{1em}

                \textbf{Ground Truth:} Parkinsonia
                \vspace{1em}

                \textbf{Predictions:}
                \vspace{1em}
                
                \begin{tabular}{p{0.28\linewidth}|p{0.28\linewidth}|p{0.28\linewidth}}
                    \textbf{Model Name} & \textbf{0 shot} & \textbf{8 shot} \\
                    \hline
                    Gemini-pro-1.5 & nan & nan \\
                    GPT-4o & Parthenium & Negative \\
                    LLaVA v1.6 34B & nan & Snake weed \\
                    Claude-3.5-sonnet & Snake weed & Parthenium \\
                    Gemini-flash-1.5 & nan & Siam weed \\
                    Claude-3-haiku & Parthenium & Snake weed
                \end{tabular}
            \end{figure}

            \begin{figure}[t!]
                \small
                \textbf{What is the severity of yellow rust disease?}
                \vspace{1em}
                \begin{center}
                \includegraphics[width=0.80\linewidth]{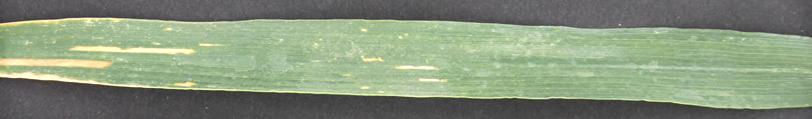}
                \end{center}
                \vspace{1em}
                
                \begin{tabular}{|p{0.28\linewidth}|p{0.28\linewidth}|p{0.28\linewidth}|}
                    \hline
                    \textbf{Category} & \textbf{Subcategory} & \textbf{Task} \\
                    \hline
                    Classification (C) & Disease Severity & Yellow Rust 19 \\
                    \hline
                \end{tabular}
                \vspace{1em}

                \textbf{Ground Truth:} MRMS
                \vspace{1em}

                \textbf{Predictions:}
                \vspace{1em}
                
                \begin{tabular}{p{0.28\linewidth}|p{0.28\linewidth}|p{0.28\linewidth}}
                    \textbf{Model Name} & \textbf{0 shot} & \textbf{8 shot} \\
                    \hline
                    Gemini-pro-1.5 & Moderately Resistant (MR) & Moderately Resistant (MR) \\
                    GPT-4o & Moderately Susceptible (MS) & Moderately Resistant (MR) \\
                    LLaVA v1.6 34B & Susceptible (S) & No disease (0) \\
                    Claude-3.5-sonnet & Moderately Resistant (MR) & No disease (0) \\
                    Gemini-flash-1.5 & Moderately Resistant (MR) & MRMS \\
                    Claude-3-haiku & Susceptible (S) & Moderately Resistant (MR)
                \end{tabular}
            \end{figure}

            \begin{figure}[t!]
                \small
                \textbf{What is the severity of yellow rust disease?}
                \vspace{1em}
                \begin{center}
                \includegraphics[width=0.80\linewidth]{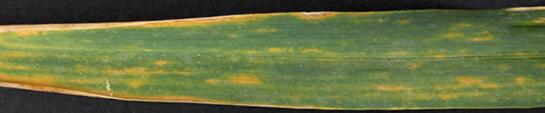}
                \end{center}
                \vspace{1em}
                
                \begin{tabular}{|p{0.28\linewidth}|p{0.28\linewidth}|p{0.28\linewidth}|}
                    \hline
                    \textbf{Category} & \textbf{Subcategory} & \textbf{Task} \\
                    \hline
                    Classification (C) & Disease Severity & Yellow Rust 19 \\
                    \hline
                \end{tabular}
                \vspace{1em}

                \textbf{Ground Truth:} Resistant (R)
                \vspace{1em}

                \textbf{Predictions:}
                \vspace{1em}
                
                \begin{tabular}{p{0.28\linewidth}|p{0.28\linewidth}|p{0.28\linewidth}}
                    \textbf{Model Name} & \textbf{0 shot} & \textbf{8 shot} \\
                    \hline
                    Gemini-pro-1.5 & Moderately Resistant (MR) & Susceptible (S) \\
                    GPT-4o & Moderately Resistant (MR) & Moderately Susceptible (MS) \\
                    LLaVA v1.6 34B & Susceptible (S) & Moderately Susceptible (MS) \\
                    Claude-3.5-sonnet & Moderately Susceptible (MS) & Moderately Susceptible (MS) \\
                    Gemini-flash-1.5 & Moderately Resistant (MR) & Moderately Susceptible (MS) \\
                    Claude-3-haiku & Moderately Susceptible (MS) & MRMS
                \end{tabular}
            \end{figure}

            \begin{figure}[t!]
                \small
                \textbf{What is the rating (1-5) of soybean stress severity?}
                \vspace{1em}
                \begin{center}
                \includegraphics[height=0.45\linewidth]{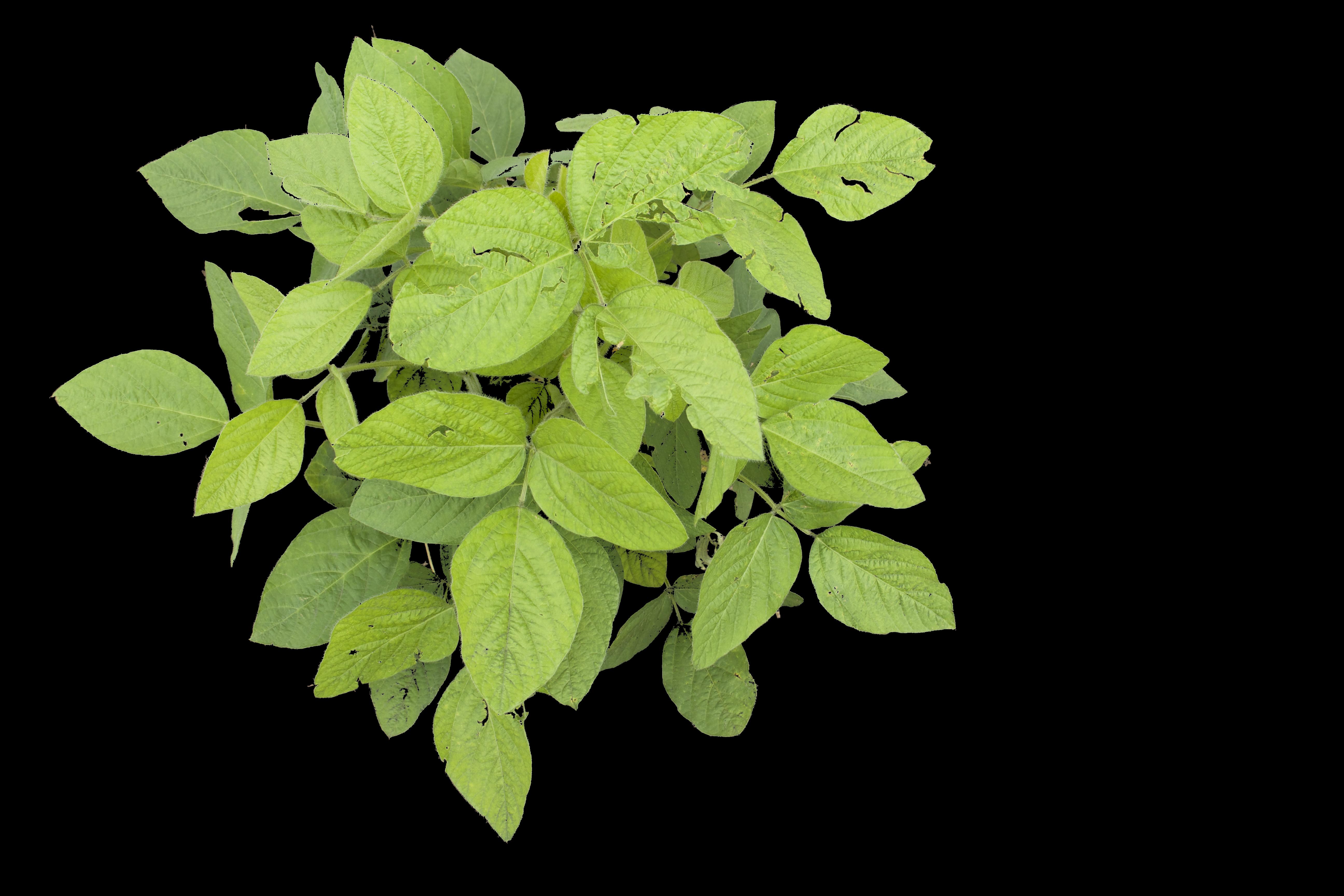}
                \end{center}
                \vspace{1em}
                
                \begin{tabular}{|p{0.28\linewidth}|p{0.28\linewidth}|p{0.28\linewidth}|}
                    \hline
                    \textbf{Category} & \textbf{Subcategory} & \textbf{Task} \\
                    \hline
                    Classification (C) & Stress Tolerance & IDC \\
                    \hline
                \end{tabular}
                \vspace{1em}

                \textbf{Ground Truth:} 1
                \vspace{1em}

                \textbf{Predictions:}
                \vspace{1em}
                
                \begin{tabular}{p{0.28\linewidth}|p{0.28\linewidth}|p{0.28\linewidth}}
                    \textbf{Model Name} & \textbf{0 shot} & \textbf{8 shot} \\
                    \hline
                    Gemini-pro-1.5 & 1.0 & 1.0 \\
                    GPT-4o & 3 & 1 \\
                    LLaVA v1.6 34B & 4.0 & nan \\
                    Claude-3.5-sonnet & 2.0 & 2 \\
                    Gemini-flash-1.5 & 1 & 2 \\
                    Claude-3-haiku & 1.0 & 3.0
                \end{tabular}
            \end{figure}

            \begin{figure}[t!]
                \small
                \textbf{What is the rating (1-5) of soybean stress severity?}
                \vspace{1em}
                \begin{center}
                \includegraphics[height=0.45\linewidth]{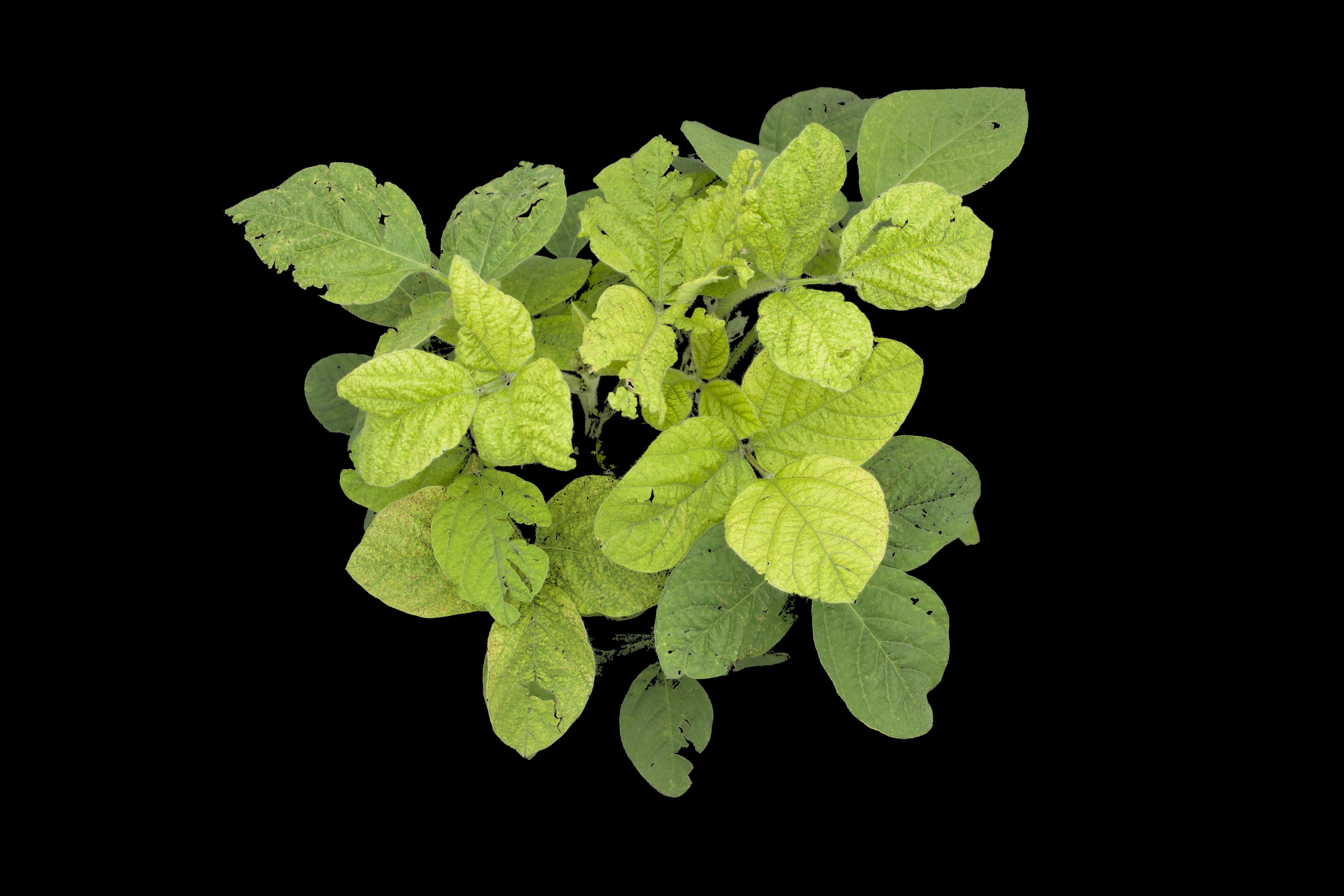}
                \end{center}
                \vspace{1em}
                
                \begin{tabular}{|p{0.28\linewidth}|p{0.28\linewidth}|p{0.28\linewidth}|}
                    \hline
                    \textbf{Category} & \textbf{Subcategory} & \textbf{Task} \\
                    \hline
                    Classification (C) & Stress Tolerance & IDC \\
                    \hline
                \end{tabular}
                \vspace{1em}

                \textbf{Ground Truth:} 2
                \vspace{1em}

                \textbf{Predictions:}
                \vspace{1em}
                
                \begin{tabular}{p{0.28\linewidth}|p{0.28\linewidth}|p{0.28\linewidth}}
                    \textbf{Model Name} & \textbf{0 shot} & \textbf{8 shot} \\
                    \hline
                    Gemini-pro-1.5 & 2.0 & 4.0 \\
                    GPT-4o & 4 & 2 \\
                    LLaVA v1.6 34B & 3.0 & nan \\
                    Claude-3.5-sonnet & 3.0 & 3 \\
                    Gemini-flash-1.5 & 2 & 3 \\
                    Claude-3-haiku & 1.0 & 3.0
                \end{tabular}
            \end{figure}

            \begin{figure}[t!]
                \small
                \textbf{What is the severity of chickpea fusarium wilt?}
                \vspace{1em}
                \begin{center}
                \includegraphics[height=0.45\linewidth]{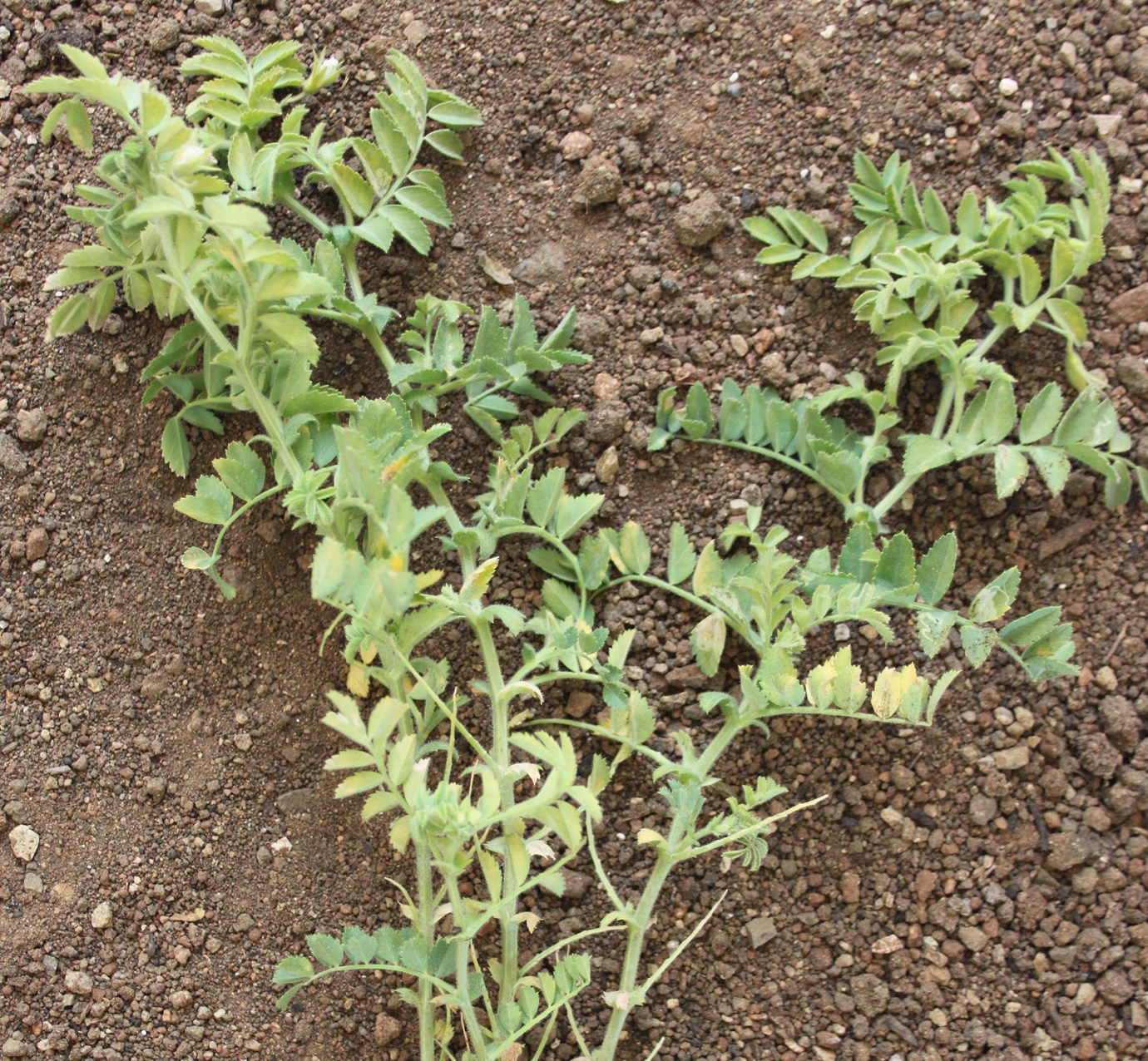}
                \end{center}
                \vspace{1em}
                
                \begin{tabular}{|p{0.28\linewidth}|p{0.28\linewidth}|p{0.28\linewidth}|}
                    \hline
                    \textbf{Category} & \textbf{Subcategory} & \textbf{Task} \\
                    \hline
                    Classification (C) & Stress Tolerance & FUSARIUM 22 \\
                    \hline
                \end{tabular}
                \vspace{1em}

                \textbf{Ground Truth:} Resistant
                \vspace{1em}

                \textbf{Predictions:}
                \vspace{1em}
                
                \begin{tabular}{p{0.28\linewidth}|p{0.28\linewidth}|p{0.28\linewidth}}
                    \textbf{Model Name} & \textbf{0 shot} & \textbf{8 shot} \\
                    \hline
                    Gemini-pro-1.5 & Susceptible & Resistant \\
                    GPT-4o & Highly Susceptible & Highly Resistant \\
                    LLaVA v1.6 34B & Highly Susceptible & nan \\
                    Claude-3.5-sonnet & Susceptible & Moderately Resistant \\
                    Gemini-flash-1.5 & Moderately Resistant & Moderately Resistant \\
                    Claude-3-haiku & Resistant & Resistant
                \end{tabular}
            \end{figure}

            \begin{figure}[t!]
                \small
                \textbf{What is the severity of chickpea fusarium wilt?}
                \vspace{1em}
                \begin{center}
                \includegraphics[height=0.45\linewidth]{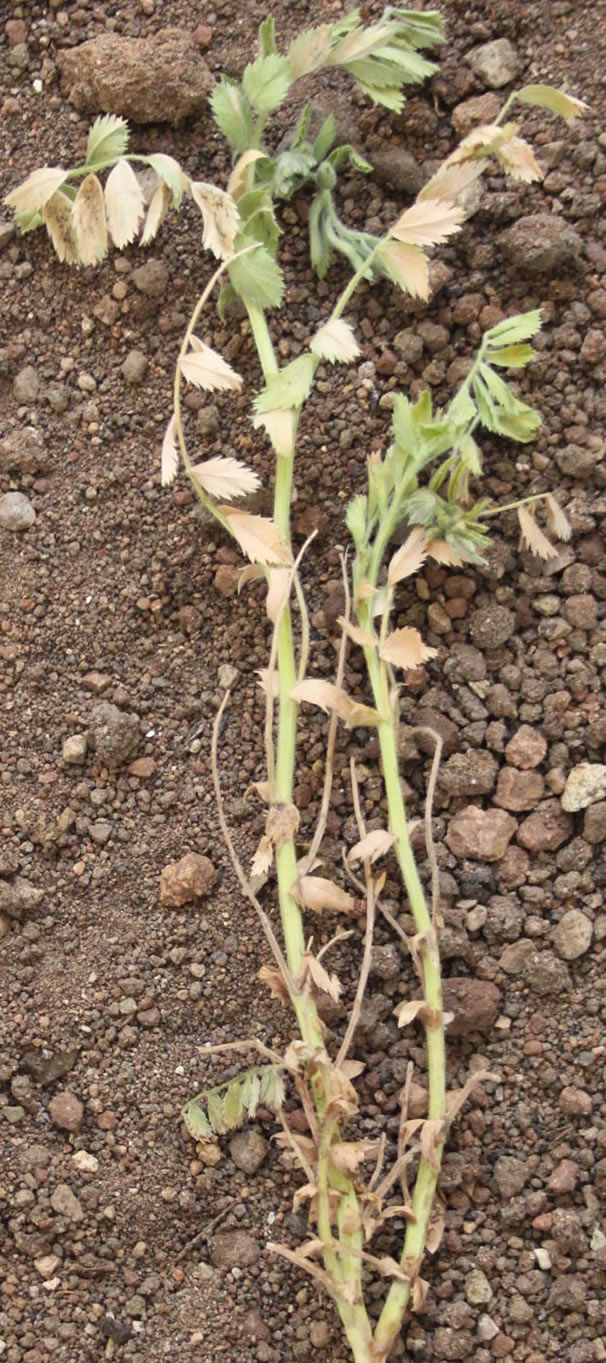}
                \end{center}
                \vspace{1em}
                
                \begin{tabular}{|p{0.28\linewidth}|p{0.28\linewidth}|p{0.28\linewidth}|}
                    \hline
                    \textbf{Category} & \textbf{Subcategory} & \textbf{Task} \\
                    \hline
                    Classification (C) & Stress Tolerance & FUSARIUM 22 \\
                    \hline
                \end{tabular}
                \vspace{1em}

                \textbf{Ground Truth:} Susceptible
                \vspace{1em}

                \textbf{Predictions:}
                \vspace{1em}
                
                \begin{tabular}{p{0.28\linewidth}|p{0.28\linewidth}|p{0.28\linewidth}}
                    \textbf{Model Name} & \textbf{0 shot} & \textbf{8 shot} \\
                    \hline
                    Gemini-pro-1.5 & Susceptible & Susceptible \\
                    GPT-4o & Highly Susceptible & Highly Susceptible \\
                    LLaVA v1.6 34B & Resistant & nan \\
                    Claude-3.5-sonnet & Susceptible & Highly Susceptible \\
                    Gemini-flash-1.5 & Highly Susceptible & Highly Susceptible \\
                    Claude-3-haiku & Susceptible & Moderately Resistant
                \end{tabular}
            \end{figure}

            \begin{figure}[t!]
                \small
                \textbf{What is the insect count?}
                \vspace{1em}
                \begin{center}
                \includegraphics[height=0.45\linewidth]{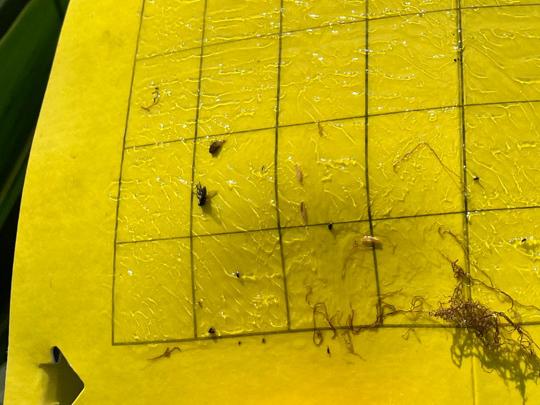}
                \end{center}
                \vspace{1em}
                
                \begin{tabular}{|p{0.28\linewidth}|p{0.28\linewidth}|p{0.28\linewidth}|}
                    \hline
                    \textbf{Category} & \textbf{Subcategory} & \textbf{Task} \\
                    \hline
                    Quantification (Q) & Pest & InsectCount \\
                    \hline
                \end{tabular}
                \vspace{1em}

                \textbf{Ground Truth:} 2
                \vspace{1em}

                \textbf{Predictions:}
                \vspace{1em}
                
                \begin{tabular}{p{0.28\linewidth}|p{0.28\linewidth}|p{0.28\linewidth}}
                    \textbf{Model Name} & \textbf{0 shot} & \textbf{8 shot} \\
                    \hline
                    Gemini-pro-1.5 & 10 & 5.0 \\
                    GPT-4o & 6 & 9 \\
                    LLaVA v1.6 34B & 7.0 & 4.0 \\
                    Claude-3.5-sonnet & 8 & 3 \\
                    Gemini-flash-1.5 & 4 & 6 \\
                    Claude-3-haiku & 17.0 & 17.0
                \end{tabular}
            \end{figure}

            \begin{figure}[t!]
                \small
                \textbf{What is the insect count?}
                \vspace{1em}
                \begin{center}
                \includegraphics[height=0.45\linewidth]{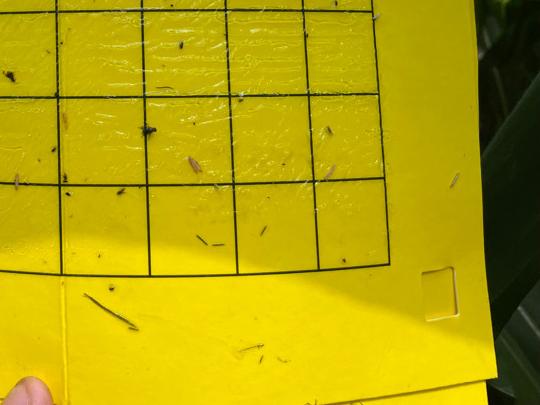}
                \end{center}
                \vspace{1em}
                
                \begin{tabular}{|p{0.28\linewidth}|p{0.28\linewidth}|p{0.28\linewidth}|}
                    \hline
                    \textbf{Category} & \textbf{Subcategory} & \textbf{Task} \\
                    \hline
                    Quantification (Q) & Pest & InsectCount \\
                    \hline
                \end{tabular}
                \vspace{1em}

                \textbf{Ground Truth:} 1
                \vspace{1em}

                \textbf{Predictions:}
                \vspace{1em}
                
                \begin{tabular}{p{0.28\linewidth}|p{0.28\linewidth}|p{0.28\linewidth}}
                    \textbf{Model Name} & \textbf{0 shot} & \textbf{8 shot} \\
                    \hline
                    Gemini-pro-1.5 & 8 & 8.0 \\
                    GPT-4o & 9 & 2 \\
                    LLaVA v1.6 34B & 0.0 & 11.0 \\
                    Claude-3.5-sonnet & 15 & 0 \\
                    Gemini-flash-1.5 & 1 & 2 \\
                    Claude-3-haiku & 22.0 & 3.0
                \end{tabular}
            \end{figure}

            \begin{figure}[t!]
                \small
                \textbf{What is the diseased leaf percentage?}
                \vspace{1em}
                \begin{center}
                \includegraphics[height=0.45\linewidth]{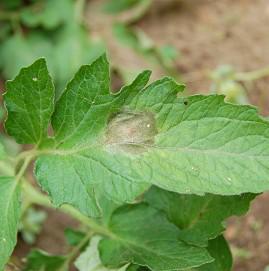}
                \end{center}
                \vspace{1em}
                
                \begin{tabular}{|p{0.28\linewidth}|p{0.28\linewidth}|p{0.28\linewidth}|}
                    \hline
                    \textbf{Category} & \textbf{Subcategory} & \textbf{Task} \\
                    \hline
                    Quantification (Q) & Disease & PlantDoc \\
                    \hline
                \end{tabular}
                \vspace{1em}

                \textbf{Ground Truth:} 3
                \vspace{1em}

                \textbf{Predictions:}
                \vspace{1em}
                
                \begin{tabular}{p{0.28\linewidth}|p{0.28\linewidth}|p{0.28\linewidth}}
                    \textbf{Model Name} & \textbf{0 shot} & \textbf{8 shot} \\
                    \hline
                    Gemini-pro-1.5 & 10.0 & 3.0 \\
                    GPT-4o & 7 & 8 \\
                    LLaVA v1.6 34B & 5.0 & 7.0 \\
                    Claude-3.5-sonnet & 12 & 4 \\
                    Gemini-flash-1.5 & 5 & 4 \\
                    Claude-3-haiku & 19.0 & 3.0
                \end{tabular}
            \end{figure}

            \begin{figure}[t!]
                \small
                \textbf{What is the diseased leaf percentage?}
                \vspace{1em}
                \begin{center}
                \includegraphics[height=0.45\linewidth]{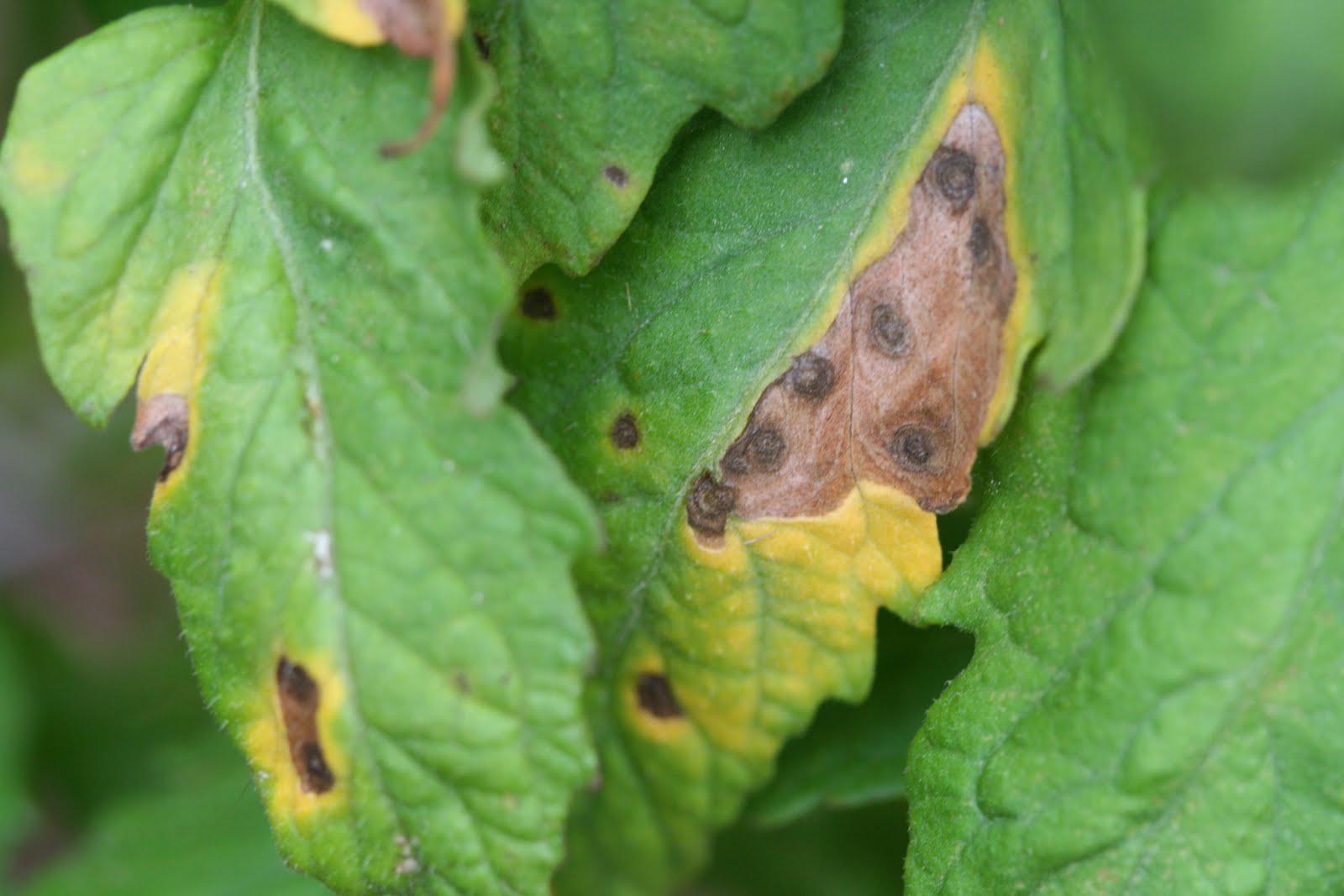}
                \end{center}
                \vspace{1em}
                
                \begin{tabular}{|p{0.28\linewidth}|p{0.28\linewidth}|p{0.28\linewidth}|}
                    \hline
                    \textbf{Category} & \textbf{Subcategory} & \textbf{Task} \\
                    \hline
                    Quantification (Q) & Disease & PlantDoc \\
                    \hline
                \end{tabular}
                \vspace{1em}

                \textbf{Ground Truth:} 12
                \vspace{1em}

                \textbf{Predictions:}
                \vspace{1em}
                
                \begin{tabular}{p{0.28\linewidth}|p{0.28\linewidth}|p{0.28\linewidth}}
                    \textbf{Model Name} & \textbf{0 shot} & \textbf{8 shot} \\
                    \hline
                    Gemini-pro-1.5 & 10.0 & 5.0 \\
                    GPT-4o & 18 & 12 \\
                    LLaVA v1.6 34B & 10.0 & nan \\
                    Claude-3.5-sonnet & 23 & 15 \\
                    Gemini-flash-1.5 & 32 & 12 \\
                    Claude-3-haiku & 18.0 & 30.0
                \end{tabular}
            \end{figure}

\end{document}